\PassOptionsToPackage{numbers}{natbib}
\documentclass{article}

\usepackage[final]{neurips_2025}

\usepackage[utf8]{inputenc} 
\usepackage[T1]{fontenc}    
\usepackage{hyperref}       
\usepackage{url}            
\usepackage{booktabs}       
\usepackage{amsfonts}       
\usepackage{nicefrac}       
\usepackage{microtype}      
\usepackage{xcolor}         
\usepackage{microtype}
\usepackage{graphicx}
\usepackage{subfigure}
\usepackage{tabularx} 
\usepackage{booktabs} 
\usepackage{xcolor}
\usepackage{amsmath}
\usepackage{amssymb}
\usepackage{mathtools}
\usepackage{amsthm}
\usepackage{algorithm}
\usepackage{algorithmic}
\usepackage{float}

\usepackage{wrapfig}

\theoremstyle{plain}
\newtheorem{theorem}{Theorem}

\newtheorem{lemma}{Lemma}      

\theoremstyle{definition}

\theoremstyle{remark}

\usepackage{caption}
\title{Information-Theoretic Reward Decomposition for Generalizable RLHF}

\author{
    Liyuan Mao\textsuperscript{1}\thanks{Work done during the internship at Institute of Artificial Intelligence (TeleAI), China Telecom.}\,\,\,, Haoran Xu\textsuperscript{2}, Amy Zhang\textsuperscript{2}, Weinan Zhang\textsuperscript{1}\textsuperscript{$\dagger$}, Chengjia Bai\textsuperscript{3}\thanks{Corresponding Author.}\\
    $^{1}$Shanghai Jiao Tong University, $^{2}$UT Austin, $^{3}$Institute of Artificial Intelligence, China Telecom \\
    \texttt{maoliyuan@sjtu.edu.cn, wnzhang@sjtu.edu.cn, baicj@chinatelecom.cn}
}

\begin{document}

\maketitle

\begin{abstract}
A generalizable reward model is crucial in Reinforcement Learning from Human Feedback (RLHF) as it enables correctly evaluating unseen prompt-response pairs.
However, existing reward models can lack this ability, as they are typically trained by increasing the reward gap between the chosen and rejected responses, while overlooking the prompts that the responses are conditioned on. Consequently, when the trained reward model is evaluated on prompt-response pairs that lie outside the data distribution, neglecting the effect of prompts may result in poor generalization of the reward model. To address this issue, we decompose the reward value into two independent components: prompt-free reward and prompt-related reward. Prompt-free reward represents the evaluation that is determined only by responses, while the prompt-related reward reflects the reward that derives from both the prompt and the response. 
We extract these two components from an information-theoretic
perspective, which requires no extra models.
Subsequently, we propose a new reward learning algorithm by prioritizing data samples based on their prompt-free reward values. 
Through toy examples, we demonstrate that the extracted prompt-free and prompt-related rewards effectively characterize the two parts of the reward value. Further, standard evaluations show that our method improves both the alignment performance and the generalization capability of the reward model.
\end{abstract}

\section{Introduction}
Reinforcement Learning from Human Feedback (RLHF) is an effective approach for Large Language Models (LLMs) alignment \cite{christiano2017deep, bai2022training}. Within a wide range of RLHF methods, reward learning plays a pivotal role. These methods typically first train a reward model on a static dataset and then leverage it to do Reinforcement Learning (RL) \cite{ouyang2022training, dong2024rlhf}. Compared with methods that are free of using reward models \cite{rafailov2024direct, zhao2023slic}, the advantage of such methods is their capacity to leverage the generalization capability of the reward model to evaluate outside-of-distribution prompt-response pairs. These prompt-response pairs with generated rewards can be used to further improve the LLM's performance \cite{stiennon2020learning, ziegler2019fine}.  

Clearly, learning a generalizable reward model is central to this scenario. However, we found that standard reward training does not guarantee sufficient generalization capability. In reward model training, the primary goal is typically better distinguishing between chosen and rejected responses. To achieve this, the reward model does not necessarily require consideration of the corresponding prompt. Taking reward learning based on Bradley-Terry (BT) model as an example. 
Since the potential response space is vastly larger than the dataset size, different data samples typically contain distinct response pairs. As long as the reward gap within each response pair increases, the training loss will decrease effectively. 
This can occur even if the reward model only considers the responses and totally ignores the prompts. In this case, the trained reward model loses its generalization capability over different prompts and may exhibit incorrect preference for novel prompt-response pairs.

\begin{figure}[t] 
    \centering
    \begin{minipage}{0.46\textwidth} 
        \centering
        \includegraphics[width=\textwidth]{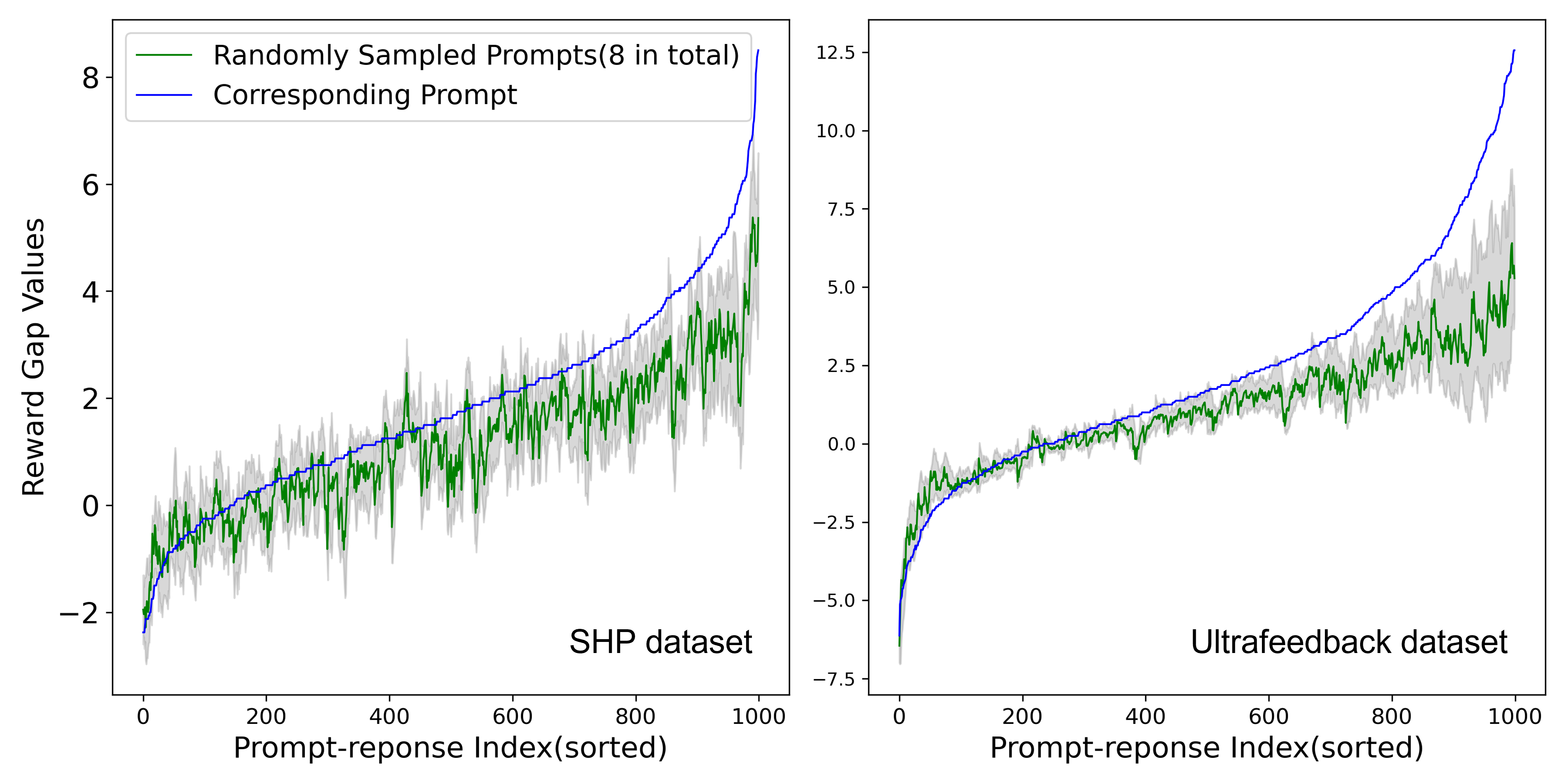} 
        \label{fig: comparison of standard reward gap and reward gaps on other prompts}
    \end{minipage}\hfill
    \begin{minipage}{0.48\textwidth} 
        \centering
        \includegraphics[width=\textwidth]{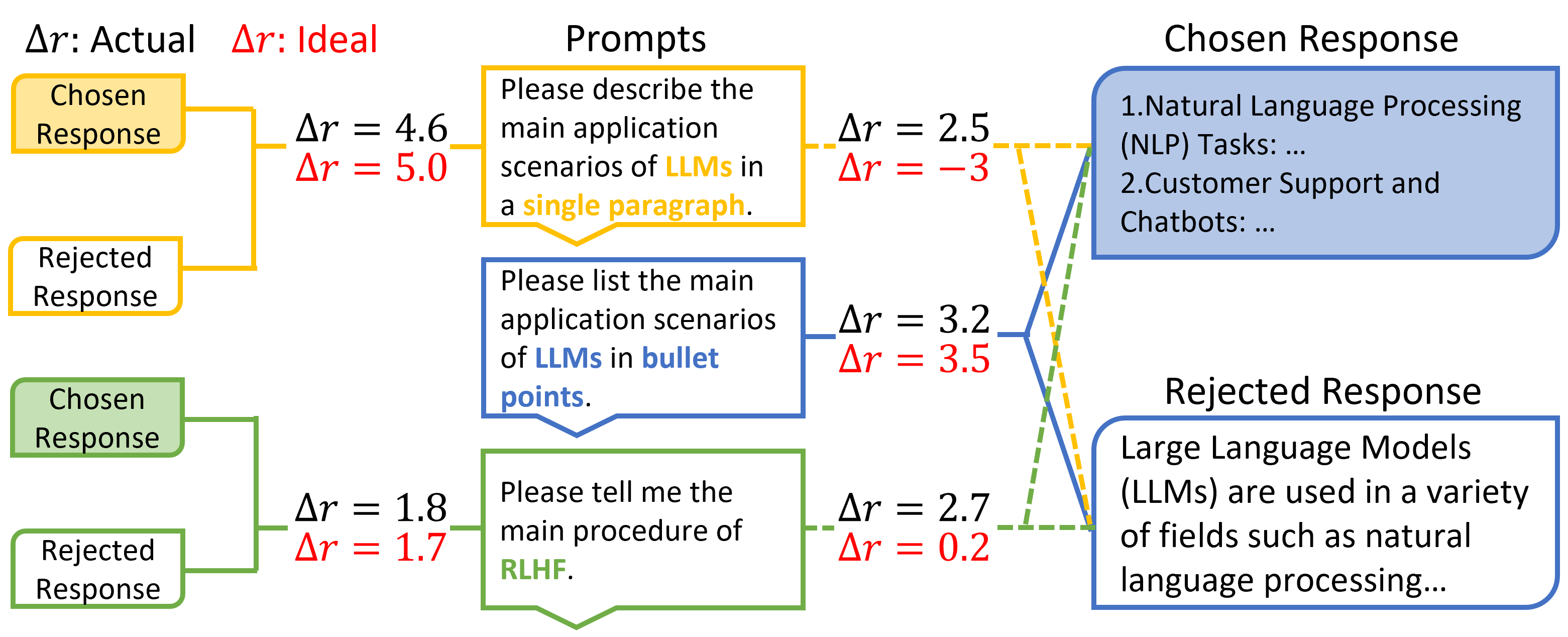} 
        \label{fig: demonstration of the catastrophic results of reward being dominated by responses}
    \end{minipage}
    \vspace{-15pt}
    \caption[Reward gap analysis on training datasets]{
\textbf{Left}: reward gaps calculated with corresponding prompt and randomly sampled prompts using QRM-Llama3-8B\footnotemark[1] on two different datasets that were used for training. When calculating with other prompts, the curves show the mean and the std. 
\textbf{Right}: illustrative failure case where the reward gap overly depends on the responses. Solid lines represent corresponding prompt-response pairs (used in training), while dashed lines represent non-corresponding pairs (unseen during training). Since the reward gap overly depends on the responses, it generalizes poorly to novel prompt-response pairs constructed even with seen prompts.
}
\label{fig: intro-fig}
\vspace{-1em}
\end{figure}

Perhaps surprisingly, such a phenomenon indeed appears in current reward models, even some that achieve SOTA performance on common benchmarks. 
As shown in Fig.~\ref{fig: intro-fig} (left), after replacing the corresponding prompt with other prompts in the dataset, the reward gaps still center around their original values. 
This issue, where responses dominate the reward gap, does not affect training but leads to catastrophic results when evaluating novel prompt-response pairs. The illustrative example in Fig.~\ref{fig: intro-fig} (right) shows this. When considering each prompt-response pair separately within the training dataset, its reward gap matches the ideal value.
However, when querying preferences after replacing the original prompt with other prompts in the dataset (which are also meaningful queries), the reward model can yield inaccurate or even wrong preferences. This generalization issue will become more pronounced when dealing with unseen prompt-response pairs encountered during evaluation. 
All of these highlight the need to distinguish two components of the reward value: one part is the value determined solely by the response, and the other is the value that can only be derived by simultaneously considering both the prompt and the response. We refer to the former as \emph{prompt-free} reward and the latter as \emph{prompt-related} reward.

To address this, we propose a novel method of decomposition to extract these two components from an information-theoretic perspective, without requiring extra models.
After that, we use the extracted prompt-free reward to guide the reward learning process, prioritizing training samples based on their prompt-free reward gap values. 
We verify our method through several toy examples and standard evaluations based on commonly used datasets and base models. 
In toy examples, the extracted prompt-free reward gaps reflect the reward model’s preference bias about response-only features, while the prompt-related reward gaps capture its generalizable preference information.
Moreover, in standard experiments with common datasets, the reward model trained with our method outperforms strong baselines. These experiments show that considering both prompt-free and prompt-related rewards during training enhances the alignment performance and generalization capabilities of the reward model.

\setcounter{footnote}{1}
\footnotetext{\url{https://huggingface.co/nicolinho/QRM-Llama3-8B}}

\section{Preliminaries}
Standard preference learning assumes the existence of the preference oracle which determines $\mathbb{P}(y_w\succ y_l|x)$, the probability that response $y_w$ is more preferred than $y_l$ conditioned on the prompt $x$. Given a preference dataset $D=\{ (x, y_w, y_l)_i \}_{i=1}^N$ where the prompt-response tuples $(x, y_w, y_l)$ are generated following $\mathbb{P}(y_w\succ y_l|x)$, our objective is to estimate the preference oracle from it. 

There exist several methods to model the preference oracle \cite{luce1959individual}, among which the Bradley-Terry (BT) model \cite{bradley1952rank} is the most widely used. The BT model further assumes that the preference oracle can be represented as a reward model $r: (x, y) \to \mathbb{R}$ that satisfies:
\begin{equation}
    \mathbb{P}_{r}(y_w\succ y_l|x)=\frac{\exp(r(x, y_w))}{\exp(r(x, y_w)) + \exp(r(x, y_l))}.
\end{equation}
Under this assumption, standard methods leverage a parameterized reward model $r_\theta$ to perform maximum likelihood estimation on the preference dataset \cite{ouyang2022training, stiennon2020learning} via the following objective:
\begin{equation}
    \max \mathbb{E}_{(x, y_w, y_l)\sim D}[\log \mathbb{P}_{r_\theta}(y_w\succ y_l|x)].
\end{equation}
In this work, we consider the preference learning based on the BT model, owing to its widespread use and strong performance in the field of RLHF.
The parameterized reward model $r_\theta$ can be implemented in various ways \cite{touvron2023llama, dong2024rlhf, wang2024helpsteer2, ivison2023camels}. 
Among these, the most commonly used one is the `Sequence Classifier' \cite{lambert2024rewardbench, ouyang2022training}, which builds the reward model on top of an LLM backbone. By projecting the feature representation of the prompt-response sequence to a scalar value using a simple network (e.g., single linear layer), such a reward model effectively leverages the prior knowledge embedded in the LLM backbone to evaluate prompt-response pairs. 
In this work, we focus on reward models within this type. A discussion of other reward model structures is given in Appendix \ref{app: Discussion of other reward modeling methods}.

\section{Methods}
\label{section: method}

In this section, we first present a specific form of decomposition that divides the reward value into a prompt-free reward and a prompt-related reward. We then formalize the problem of such decomposition and propose a solution from an MI perspective. Finally, we use the extracted prompt-free rewards to prioritize training samples and guide reward learning.

\subsection{Prompt-free Reward \& Prompt-related Reward}
\label{subsec: problem definition}

We decompose the reward value $r_\theta(x, y)$ into two separate parts: the prompt-free reward and the prompt-related reward. The prompt-free reward is only determined by the response and can be regarded as the overall evaluation of the response. Specifically, the prompt-free reward ($r_2$) satisfies Eq. \eqref{eq: property of prompt-free reward}. Once the response is given, the prompt-free reward remains unaffected by any specific prompt.
\begin{equation}
\label{eq: property of prompt-free reward}
    \forall x_1, x_2, y ~~\in D , ~ r_2(x_1, y)=r_2(x_2, y).
\end{equation}
In contrast, as the other part of $r_\theta(x, y)$ besides the prompt-free reward, prompt-related reward ($r_1$) varies among different prompts. Such a reward can be determined only when both the prompt and the response are given. 

Given a reward model $r_\theta$ during training and a prompt-response tuple $(x, y_1, y_2)$ sampled from the dataset, our goal is to identify the prompt-free reward in $r_\theta(x, y_1)$ and $r_\theta(x, y_2)$ and leverage it to guide reward learning. 
We consider decomposing $r_\theta(x, y)$ into the additive form: 
\begin{equation}
    r_\theta(x, y) = r_1(x, y) + r_2(x, y).
\end{equation}
Examining the dataset holistically, $r_2(x, y)$ demonstrates clear randomness, with the data distribution influencing its value.
Specifically, for a given $(x, y)$ pair, if the prompts related to $y$ are diverse, there is little connection between $r_2(x, y)$ and $r_\theta(x, y)$, since $r_2$ remains unchanged across different prompts, whereas $r_\theta$ changes accordingly.
Conversely, if the related prompts show little variety, $r_2(x, y)$ can be closely related to $r_\theta(x, y)$, since $r_\theta(x, y)$ also exhibits little change.
This motivates us to define $r_2$ by considering the randomness in the data distribution, treating $r_2$ as the joint product of $r_\theta$ and the data distribution. We also note that although some simple definitions of $r_2$ may seem reasonable (e.g. $r_2(x, y) = \mathbb{E}_{x' \sim P(X|Y=y)}[r_\theta(x', y)]$), they can't truly reflect $r_\theta$'s preference regarding the response alone (See Appendix \ref{app: Numerical Example} for more details).



\subsection{Solving Prompt-free Reward via a Mutual Information (MI) Objective}
\label{subsec: Solving Context-free Reward via a MI Objective}
In this section, we extract \( r_2(x, y) \) from \( r_\theta(x, y) \) via a preference perspective, with a carefully designed mutual information (MI) objective. A brief introduction to MI is provided in Appendix \ref{app: Brief Introduction to Mutual Information}. For simplicity, we have notations
\begin{equation}
    \Delta r(x, y_1, y_2) \coloneqq r(x, y_1) - r(x, y_2) \quad \sigma(\Delta r(x, y_1, y_2)) \coloneqq 1/\big(1 + \exp(-\Delta r(x, y_1, y_2))\big)
\end{equation}
Since we focus on preference learning based on the BT model, 
it is sufficient to study the reward gap ($\Delta r$) rather the reward itself, as the reward gap decides the preference label. Consequently, we decompose the reward gap as 
\begin{equation}
\Delta r_{\theta}(x,y_1,y_2)=\Delta r_1(x,y_1,y_2)+\Delta r_2(x,y_1,y_2).
\end{equation}
Meanwhile, since both \( r_1 \) and \( r_2 \) must be applicable to all \( (x, y_1, y_2) \), it is essential to take into account the entire dataset when decomposing from the preference perspective. The preference labels of \( r_1 \), \( r_2 \), and \( r_\theta \) over the entire dataset are inherently random and can be characterized by the following random variables:
\begin{equation}
\begin{aligned} 
Z\coloneqq\text{Ber}\big(\sigma(\Delta r_1(X,Y_1,Y_2)\big),\qquad\qquad\qquad\quad~~~&\tilde{Z}\coloneqq\text{Ber}\big(\sigma(\Delta r_2(X,Y_1,Y_2)\big)\\
\tilde{W}\coloneqq\text{Ber}(\mathbb{E}_{x\sim P(X|Y_1,Y_2)}[\sigma(\Delta r_\theta(x,Y_1,Y_2))]), \quad& W\coloneqq\text{Ber}\big(\sigma(\Delta r_\theta(X,Y_1,Y_2))\big)\\
\end{aligned}
\end{equation}
where `Ber' stands for Bernoulli random variable. Among them, $Z$, $\tilde{Z}$ and $W$ are the random preference labels of their corresponding rewards. $\tilde{W}$ is the random prompt-free preference label of $r_\theta$. For all these random variables, the randomness comes from the prompt, responses, and the Bernoulli distribution. We also provide a detailed interpretation of these random variables in Appendix \ref{app: Interpretation of the random variables}.

Based on the defined Bernoulli random variables, we proceed to characterize the desired $r_1$ and $r_2$. 
Recall that $r_1$ and $r_2$ represent prompt-related and prompt-free rewards, respectively. This means $Z$ should encapsulate only prompt-related preference while $\tilde{Z}$ should only reflect prompt-free preference. Formally, this could be written in the following conditions:
\begin{equation}
    \text{MI}(Z~\|~\tilde{W})=0, \quad \text{MI}(\tilde{Z}~\|~\tilde{W}) = \text{MI}(\tilde{Z}~\|~W).
\label{eq: conditions without entropy}
\end{equation}

\begin{wrapfigure}{R}{0.5\textwidth}
\vspace{-1em}
\centering
\includegraphics[width=0.15\textwidth]{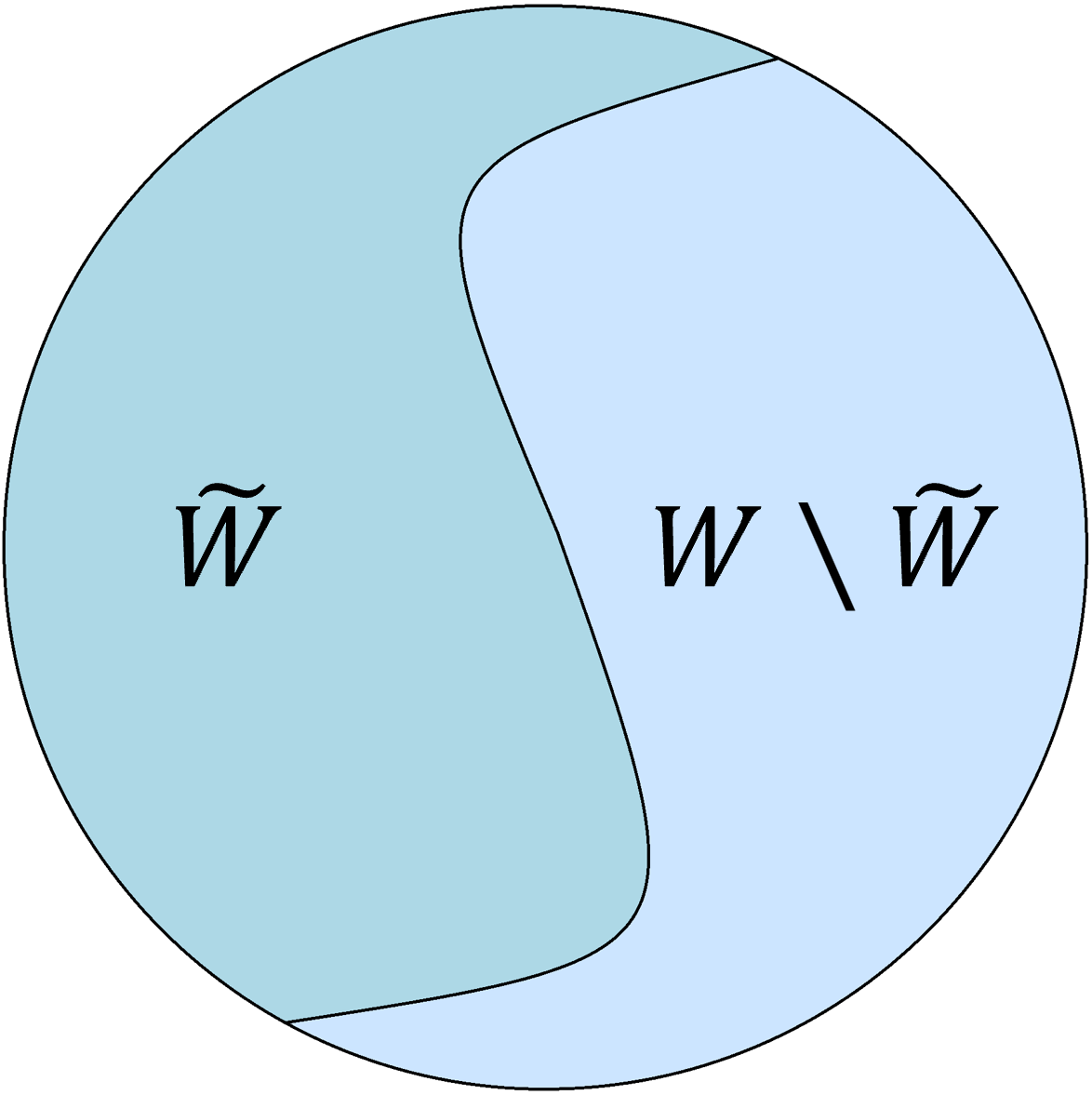}
\hspace{0.01\textwidth}
\includegraphics[width=0.15\textwidth]{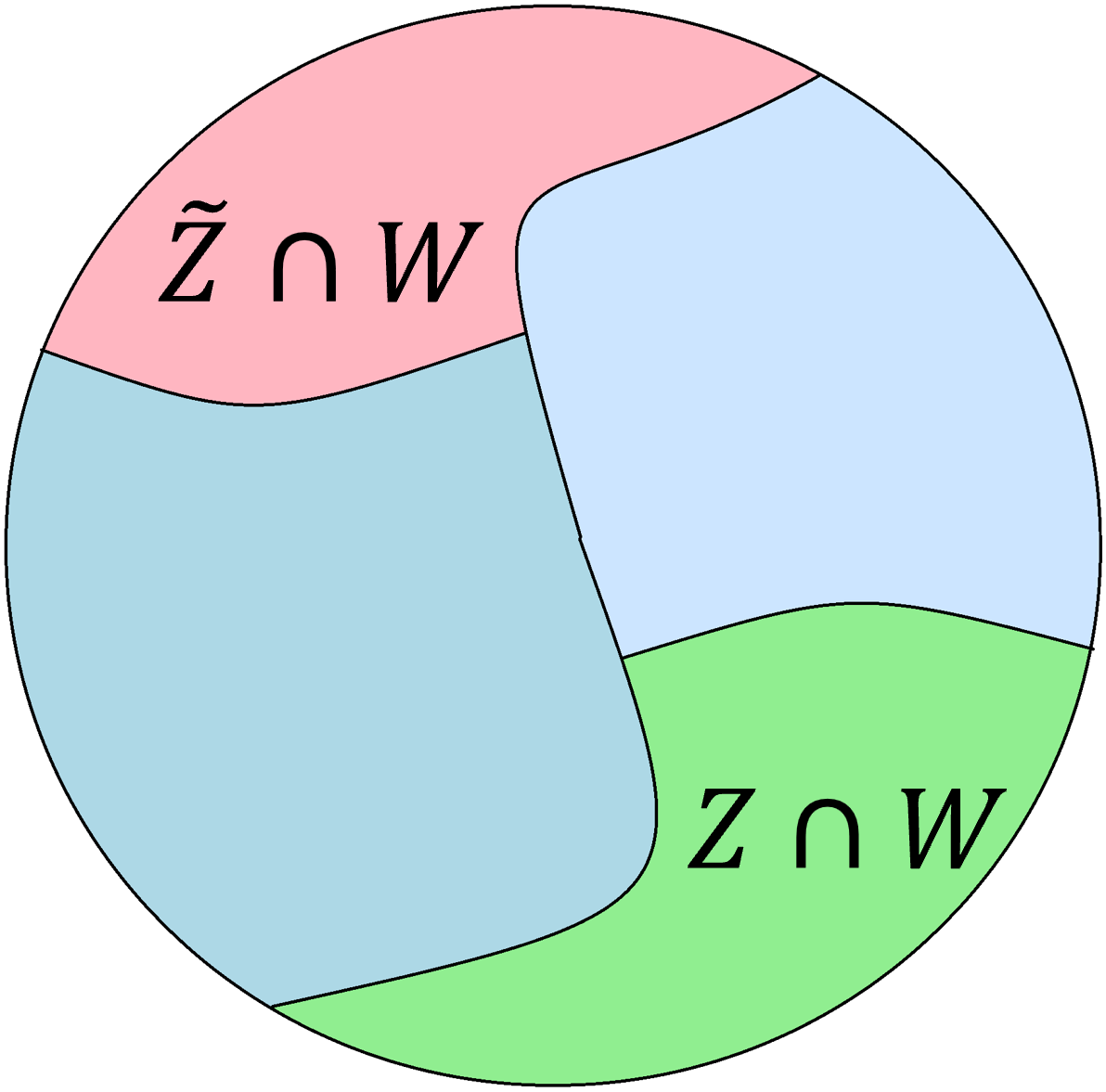}
\hspace{0.01\textwidth}
\includegraphics[width=0.15\textwidth]{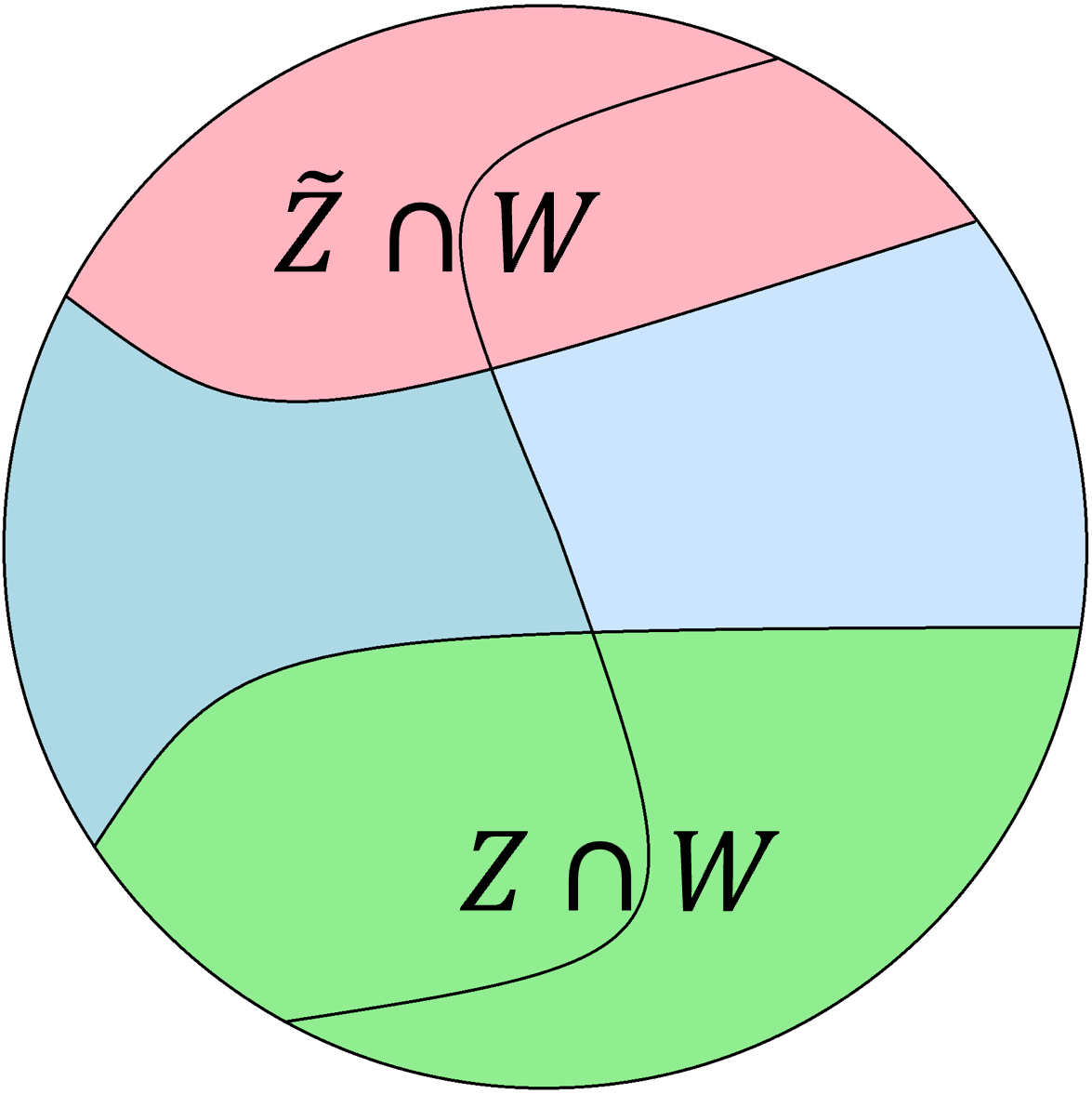}
\vspace{-0.5em}
\caption{(a) Information in $W$ and $\tilde{W}$. (b) Desired information in $Z$ and $\tilde{Z}$. (c) Undesired information in $Z$ and $\tilde{Z}$.}
\label{fig: venn diagram to illustrate the desired property of r1 and r2}
\vspace{-1em}
\end{wrapfigure}

The Venn diagram in Fig.~\ref{fig: venn diagram to illustrate the desired property of r1 and r2} further illustrates these conditions. The first condition eliminates all information in $Z$ that's related to prompt-free preference, while the second condition ensures that the information contained in $\tilde{Z}$ does not exceed the prompt-free preference.

Although these conditions effectively constrain the information in $Z$ and $\tilde{Z}$, a trivial solution exists. Assume $\Delta r_\theta(x, y)$ is bounded for any prompt $x$ and response $y$. Consider an ill-formed $\Delta r_1$ such that for any prompt $x$ and response $y$, $\Delta r_1(x, y) = +\infty$. Then by the definition that $\Delta r_\theta(x, y) = \Delta r_1(x, y) + \Delta r_2(x, y)$, $\Delta r_2(x, y) = -\infty$ for any prompt $x$ and response $y$. In this case, $Z$ will always be $1$, and $\tilde{Z}$ will always be $0$. This makes $\text{MI}(Z\| \tilde{W})=\text{MI}(\tilde{Z}\|\tilde{W})=\text{MI}(\tilde{Z}\|W)=0$ since $Z$ and $\tilde{Z}$ are both constants. Although these ill-formed $r_1$ and $r_2$ satisfy the conditions in Eq.~\eqref{eq: conditions without entropy}, they provide no information of prompt-related or prompt-free preferences. 
This ill-formed example indicates that the MI condition in Eq.~\eqref{eq: conditions without entropy} is insufficient to induce a reasonable decomposition. When $Z$ and $\tilde{Z}$ become constants, the information within them is lost, along with their MI with other random variables. 
To avoid this, we optimize the following constrained objective to derive $r_1$ and $r_2$, which ensures sufficient information in $Z$ ($H$ stands for Shannon's entropy):
\begin{equation}
\begin{aligned}
\max_{r_1} \quad H(Z),
\quad \text{s.t.} \quad 
\left\{
\begin{aligned}
\text{MI}(Z\|\tilde{W})&=0 \\
\text{MI}(\tilde{Z}\|\tilde{W}) &= \text{MI}(\tilde{Z}\|W).
\end{aligned}
\right.
\end{aligned}
\label{eq: constrained optimization problem}
\end{equation}
In addition to the previous conditions, we maximize the information within $Z$, considering that $Z$ represents the random prompt-related preference label across the entire dataset. This constrained optimization problem is difficult to solve directly, and direct solutions may require additional parameterized reward models. However, we provide an efficient algorithm that obtains $\Delta r_1(x, y_1, y_2)$ and $\Delta r_2(x, y_1, y_2)$
without additional reward model. To see this, we first note that the second constraint is satisfied with a specific structure of $r_2$. Formally, the following theorem holds:
\begin{theorem}
    When the value of $r_2$ depends only on the response, i.e. $r_2(x, y) = r_2(y)$, \textnormal{MI}$(\tilde{Z}\|\tilde{W})$ $=$ \textnormal{MI}$(\tilde{Z}\|W)$.
    \label{theorem: only input response}
\end{theorem}
The proof is given in Appendix \ref{app: proof}. Intuitively, Theorem \ref{theorem: only input response} holds since such $\tilde{Z}$ inherently does not contain any information from the prompt. Consequently, removing the prompt-related information from $W$ to $\tilde{W}$ does not reduce the MI term. This important property enables us to concentrate on the first constraint and maximize the information in $Z$, as long as $r_2$ maintains this simple structure. We continue to characterize the optimal solution of Eq.~\eqref{eq: constrained optimization problem} in the following theorem.

\begin{theorem}
    For any bounded $r_\theta$ and dataset $(X, Y_1, Y_2)$, there exist \textbf{feasible} $r_1^*$, $r_2^*$ such that $\forall (y_1, y_2) \sim P(Y_1, Y_2), \mathbb{E}_{x\sim P(X|Y_1=y_1, Y_2=y_2)}[\sigma(\Delta r_1^*(x, y_1, y_2))]=\frac{1}{2}$. Such $r_1^*$, $r_2^*$ is the optimal solution to problem \eqref{eq: constrained optimization problem}.
    \label{theorem: optimal solution}
\end{theorem}
The proof is given in Appendix \ref{app: proof}. Intuitively, this requires the prompt-free part of $r_1^*$ to equally prefer both responses in any given pair. This aligns with our desiderata, as $r_1^*$ should prefer some response only after being provided with the specific prompt. 
 Although neither of the theorems explicitly shows the connection with $r_\theta$, we note that the feasibility condition requires $\Delta r_1^*(x, y_1, y_2)+\Delta r_2^*(x, y_1, y_2)$ to equal $\Delta r_\theta^*(x, y_1, y_2)$, which encodes $r_\theta$'s information.
 Theorem \ref{theorem: optimal solution} also provides an efficient way to obtain $\Delta r_1^*(x, y_1, y_2)$ and $\Delta r_2^*(x, y_1, y_2)$. After replacing $\Delta r_1^*(x, y_1, y_2)$ with $\Delta r_\theta(x, y_1, y_2) - \Delta r_2^*(x, y_1, y_2)$ in the optimal solution in Theorem \ref{theorem: optimal solution}, $\Delta r_2^*(x, y_1, y_2)$ satisfies that for any $(y_1, y_2) \sim P(Y_1, Y_2)$:
\begin{equation}
\label{eq: expectation of shifted delta r_theta}
    \Phi(y_1, y_2)\coloneqq\underset{x \sim P(X | Y_1 = y_1, Y_2 = y_2)}{\mathbb{E}}
[\sigma(\Delta r_\theta(x, y_1, y_2) - \Delta r_2^*(y_1, y_2))]=\frac{1}{2}.
\end{equation}
Note that we replace $\Delta r_2^*(x, y_1, y_2)$ with $\Delta r_2^*(y_1, y_2)$ due to Theorem \ref{theorem: only input response}. Because of this, $\Delta r_2^*(y_1, y_2)$ will not change while taking the conditional expectation.
Moreover, it can be proved that $\Phi(y_1, y_2)$ in Eq.~\eqref{eq: expectation of shifted delta r_theta} decreases monotonically with the increase of $\Delta r_2^*(y_1, y_2)$, as shown in Appendix \ref{app: proof}. Combined with the bounded assumption of $r_\theta$, we have:
\begin{equation}
    \lim_{\Delta r_2^*(y_1, y_2) \to +\infty} \Phi(y_1, y_2) = 0, \quad \lim_{\Delta r_2^*(y_1, y_2) \to -\infty} \Phi(y_1, y_2) = 1
\end{equation}
With this, $\Delta r_2^*(y_1, y_2)$ can be obtained via binary search in a finite interval. The searched $\Delta r_2^*(y_1, y_2)$ can be interpreted as the `weighted average' of $\Delta r_\theta(x, y_1, y_2)$, based on $P(X | Y_1 = y_1, Y_2 = y_2)$. Due to limited space, we provide the pseudo-code of the binary search process in Appendix \ref{app: pseudo-code} (Alg. \ref{alg: binary search}). Note that such an algorithm doesn't require any extra parameterized reward models.

It's clear that during the binary search, the estimation of the expectation in Eq. \eqref{eq: expectation of shifted delta r_theta} requires sampling from $P(X| Y_1 = y_1, Y_2 = y_2)$.
Although sampling from it is generally challenging, we propose utilizing importance sampling by considering the decomposition with Bayes' rule:
\begin{equation}
    P(x | y_1, y_2)=P(x)P( y_1, y_2|x) ~/~ P(y_1, y_2).
\end{equation}
Here $P(x | y_1, y_2)$ is the abbreviation of $P(X=x | Y_1 = y_1, Y_2 = y_2)$, so as other probabilities. Given a $(y_1, y_2)$, we first sample from $P(X)$, the marginal distribution of prompts, and then re-weight the samples using $P( y_1, y_2|x)$. Its value represents the response generation probability and thus can be efficiently computed in some cases or approximated in others. Moreover, since $P( y_1, y_2|x)$ is independent of the trained reward model, each dataset only needs to pre-sample prompts once and compute the probabilities once. We list the sampling schemes for different cases in Appendix \ref{app: Sampling schemes}. 

\subsection{Guide Reward Learning with Prompt-free Reward}
During the training process of $r_\theta$, one may consider leveraging the reward gap value $\Delta r_\theta(x,y_1,y_2)$ to selectively train more on samples with \emph{small} reward gap (including negative ones).
The reason is that, if the reward gap for $(x, y_1, y_2)$ satisfies $\Delta r_\theta(x, y_1, y_2) \approx 0$ or $< 0$, it indicates that the current reward model $r_\theta$ struggles to distinguish between the preferred and dispreferred responses. Consequently, allocating more training budget to these samples can better align the reward model with human preference. 
Similar
methods have been studied in training LLMs \cite{lin2024rho, mindermann2022prioritized}.

\textbf{Characterize ideal $r_\theta$ with $\Delta r_1$ and $\Delta r_2$.~} However, things become different if we consider prompt-related
and prompt-free reward separately. 
During the learning process of $r_\theta$, we expect (\romannumeral1)~$r_\theta$ to learn more prompt-related preference from the dataset (large $\Delta r_1(x, y_1, y_2)$); and (\romannumeral2) $r_\theta$ to learn less prompt-free preference (small \textit{absolute} values of $\Delta r_2(x, y_1, y_2)$). 
This is because when $\Delta r_2(x,y_1,y_2)\gg 0$ or $\ll 0$, it indicates strong prejudice over the sampled responses, as $\Delta r_2$ doesn't take the specific prompt into account. This prejudice may introduce spurious reward gaps (e.g., reward gap induced by length bias)\citep{dubois2024length,singhal2024a}. 


\textbf{Prioritizing data with small $\Delta r_2$.~} Similar to the method discussed at the beginning of this section, $\Delta r_1$ and $\Delta r_2$ are important properties of $r_\theta$ and can also be leveraged to guide the training of $r_\theta$. To fulfill the desiderata above, we present a straightforward method that prioritizes preference data when training reward models, with the aim of increasing prompt-related reward gaps while constraining the value of prompt-free reward gaps. Specifically, given data samples $\{(x, y_1, y_2)_i\}_{i=1}^k$ in each iteration, we first decompose $r_\theta$ to get their prompt-free reward gaps $\{\Delta r_2(x, y_1, y_2)_i\}_{i=1}^k$ based on Alg.~\ref{alg: binary search}, and then perform update for $r_\theta$ on those samples with small prompt-free reward gaps. For samples with large prompt-free gaps, we ignore them in this iteration and reinsert them into the buffer for subsequent updates.

\textbf{Analysis of prioritization mechanism.~} Next, we provide an in-depth analysis for the proposed data prioritization mechanism. The standard update following the BT model only ensures the increase of $\Delta r_1(x, y_w, y_l) + \Delta r_2(x, y_w, y_l)$, so we cannot guarantee that either of them will necessarily increase. However, (\romannumeral1) if samples have small prompt-free reward gaps (e.g. $\Delta r_2<0$), regardless of whether their prompt-related reward gaps ($\Delta r_1$) are large or small, they should be used for updates. This is because an increase in $\Delta r_1$ suggests a better mastery of prompt-related preferences, while an increase in their $\Delta r_2$ indicates the elimination of existing prejudices, both of which are beneficial.
(\romannumeral2) Conversely, for samples with large $\Delta r_2$ (e.g., $\Delta r_2\gg0$), if their $\Delta r_1$ are also large, updating is unnecessary since the prejudice led to gradient saturation in the BT model. If their $\Delta r_1$ values are small, their $\Delta r_\theta$ values are dominated by factors unrelated to the prompt (e.g., response length). Updating $r_\theta$ on them would exacerbate this issue, hindering generalization. Fig. 3 further illustrates this mechanism. Data samples with smaller $\Delta r_2$ help achieve a better trade-off between increasing prompt-related reward gaps and constraining prompt-free reward gaps.




\begin{figure}[H] 
\label{fig: mechanism}
\centering
    \includegraphics[width=0.8\textwidth]{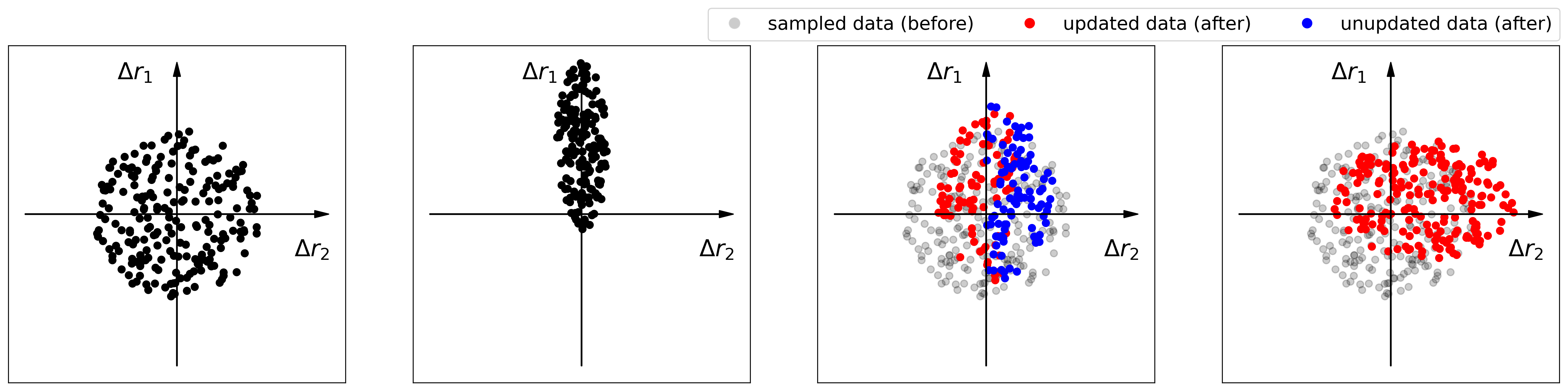}
    \vspace{-0.5em}
    \caption{(Illustrative) We characterize training data samples in a 2-dimensional quadrant diagram, with the decomposed reward gaps $\Delta r_1$ and $\Delta r_2$. (a) shows initial data samples before training. (b) After training, the ideal distribution should be centered on the positive half of the $\Delta r_1$-axis, indicating that the preference depends solely on prompt-related information. (c-d) However, in the training process of $\Delta r_\theta$, the update following the BT model can only ensure the data points move in at least one of the positive directions of the $\Delta r_1$-axis or $\Delta r_2$-axis (up or right). If the update is based on samples with small $\Delta r_2$ (e.g. the left half), their movement upwards or to the right, along with other unupdated samples (e.g. the right half), will cause the distribution to be more centered on the positive $\Delta r_1$-axis.
    On the other hand, if the update is based on all samples, as shown in (d), the movement of samples with large $\Delta r_2$ values to the right exacerbates the existing prejudices, causing the distribution to become more centered on the positive $\Delta r_2$-axis.}
    \vspace{-1.5em}
\end{figure}

\textbf{Practical Consideration.} In practice, we design a binary-cluster mechanism to determine whether the value of $\Delta r_2$ is `small' or `large' in data prioritization. Specifically, at each step, we sample a batch of \( (x, y_1, y_2) \) pairs from the preference dataset and perform one-dimensional binary clustering based on their \( \Delta r_2 \) values. Due to the dynamic nature of \( r_\theta \) and the noise in data sampling, we do not directly use the obtained decision boundary for determination. Instead, we maintain a dynamic threshold, calculated as the exponential moving average of the boundary, and use this threshold for the determination. 
Due to limited space, we provide a detailed process in Appendix \ref{app: pseudo-code} (Alg. \ref{alg: select sample with context-free reward}). We justify the use of binary clustering and the EMA threshold in Appendix \ref{app: Justification and analysis data prioritization} and demonstrate their robustness through experiments.

It is worth noting that samples with $\Delta r_2$ values exceeding the threshold will not be discarded from the dataset. Instead, the reinsertion operation assigns a probability for these samples to be updated the next time they are sampled. Since $r_\theta$ dynamically evolves during training, samples that are not prioritized at the moment may contribute significantly to the future $r_\theta$. To avoid endlessly cycling through samples with large $\Delta r_2$ values, the opportunities for reinsertion are limited. 
We provide more details of the algorithm in Appendix~\ref{app: algorithm details}.

\section{Experiments}
\label{section: experiments}

In this section, we demonstrate the significance of identifying and utilizing prompt-free rewards to guide reward learning from two perspectives. We first illustrate that, for some manually crafted datasets, the extracted prompt-free reward reflects the reward model's preference bias and the prioritization effectively aids in reward learning. Subsequently, based on some commonly used open-source preference datasets, we evaluate the trained reward model using direct metrics (reward model accuracy) and indirect metrics (performance of the induced policy). The evaluation results demonstrate that, with the guidance of prompt-free rewards, our method enhances the generalization ability of the learned reward model and further improves the performance of the induced LLM policy after alignment.


\subsection{Experiments on Manually Crafted Datasets}
\label{subsec: Experiments on Manually Crafted Datasets}

In this part, we manually construct datasets with specific characteristics from SHP \cite{pmlr-v162-ethayarajh22a}, a commonly used open-source preference dataset. Based on these datasets, we compare the reward models trained with ordinary data and prioritized data. To thoroughly demonstrate their difference 
during training, we select four equally spaced training steps, sample a number of data pairs at random, and visualize them by their
 $\Delta r_1, \Delta r_2$ values using the same method as in Fig. 3.
Additionally, we evaluate the final reward model using Reward-Bench \citep{lambert2024rewardbench} to assess its generalization capabilities. We choose LLaMA-3.2-1B-Instruct as the backbone due to its lightweight nature and strong performance. For more details of the experiments, we refer to Appendix \ref{app: experimental details}.

\subsubsection{Length-biased Dataset}
\label{subsubsec: toy examples: Length-biased Dataset}
We construct a length-biased dataset $\mathcal{D}_{\rm bias}$ that contains preference pairs with 80\% \emph{chosen-longer} responses (i.e. $|y_w| > |y_l|$) and 20\% \emph{chosen-shorter} responses. The details of the construction are given in Appendix \ref{app: experimental details}.
When performing reward learning in $\mathcal{D}_{\rm bias}$, the BT loss for a uniformly sampled data batch can be easily optimized by considering only the lengths of the responses, as the majority of the data exhibits a preference for longer chosen responses. As a result, the reward model can easily overfit to such spurious, prompt-free preferences. 

\begin{figure}[htbp]
\centering
\vspace{-0.25em}
\begin{minipage}{0.48\textwidth}
    \centering
    \includegraphics[width=\linewidth]{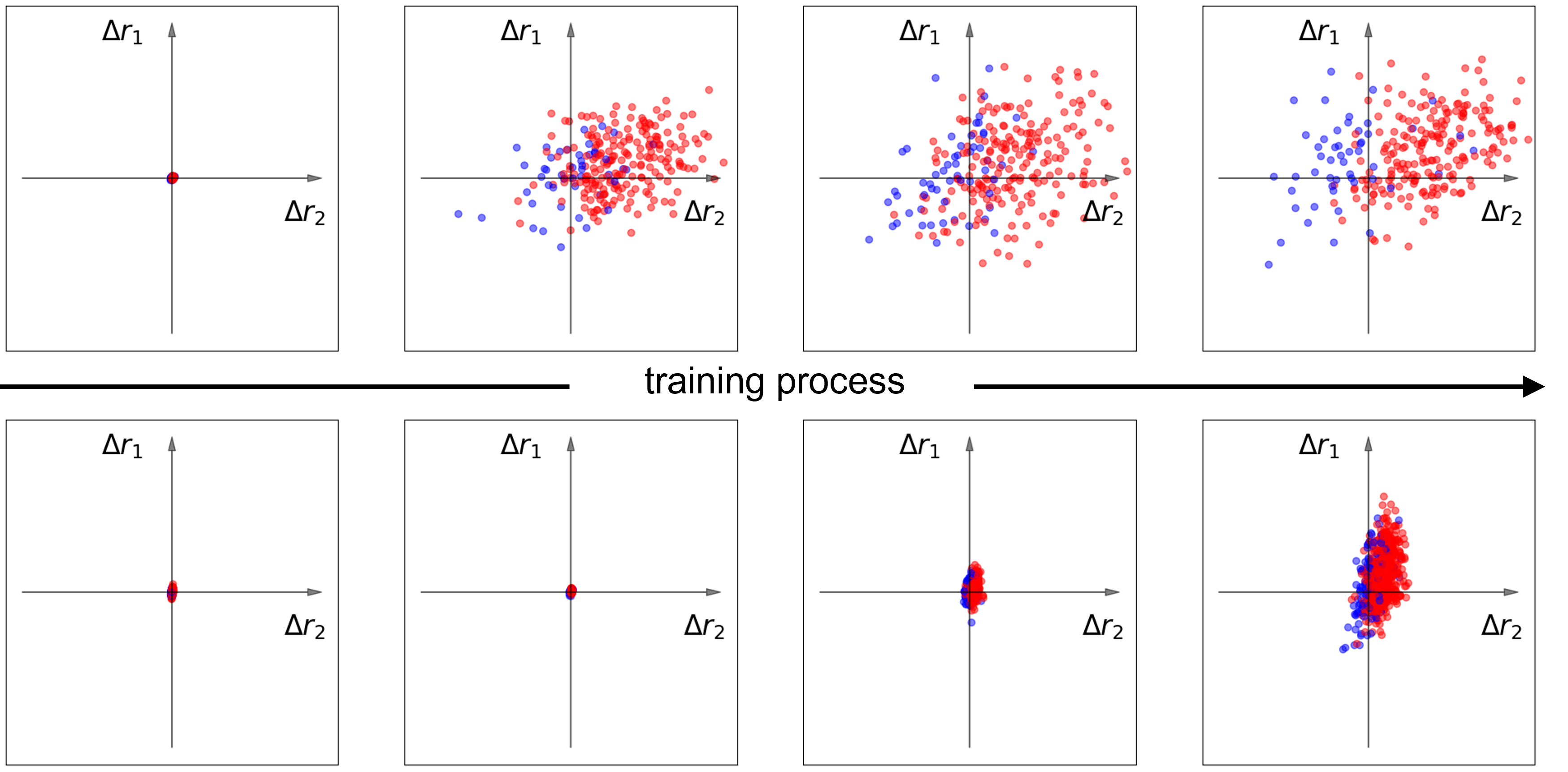}
    \small (a)
\end{minipage}
\hspace{0.02\textwidth}
\begin{minipage}{0.48\textwidth}
    \centering
    \includegraphics[width=\linewidth]{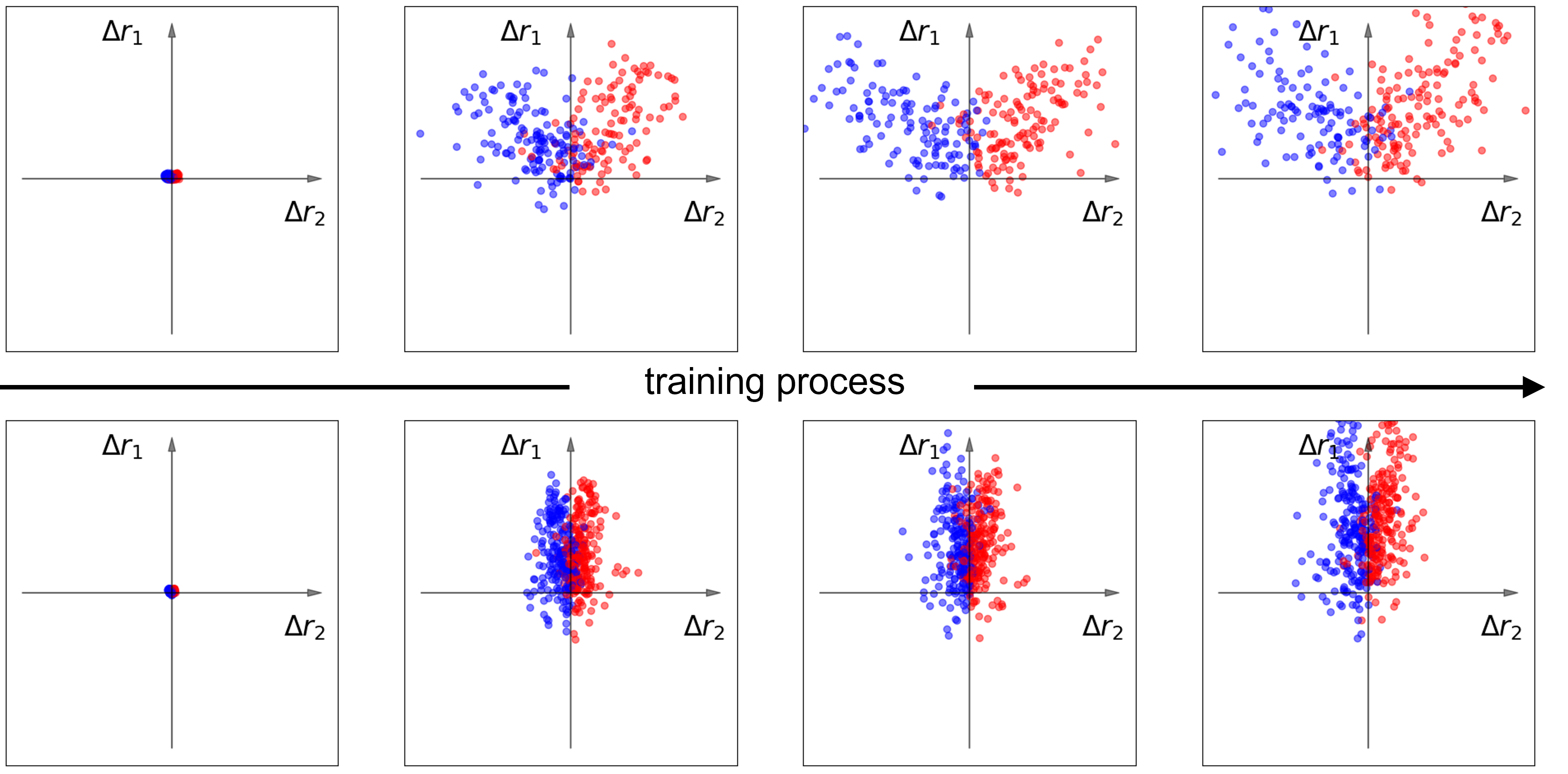}
    \small (b)
\end{minipage}
\vspace{-0.5em}
\caption{(a): Visualizations for the length-biased dataset. We mark the data points that satisfy $|y_w| > |y_l|$ in \textcolor{red}{red} and the ones that satisfy $|y_w| \leq |y_l|$ in \textcolor{blue}{blue}. (b): Visualizations for the adversarial prompt dataset. We mark adversarial data in \textcolor{red}{red} and original data in \textcolor{blue}{blue}. In both (a) and (b), the top shows ordinary training data, and the bottom shows prioritized training data.}
\label{fig:toy_examples}
\vspace{-1em}
\end{figure}

The visualizations during training and the final results on Reward-Bench are shown in Fig.~\ref{fig:toy_examples} (a) and Table \ref{tab: length-biased and adversarial} (a).
If $r_\theta$ is trained with data uniformly sampled from $\mathcal{D}_{\rm bias}$ (Fig.~\ref{fig:toy_examples} (a), top), although its prompt-related reward gap ($\Delta r_1$) increases gradually, $r_\theta$ rapidly overfit to the length preference. This results in a significantly faster increase in $\Delta r_2$ for the \emph{chosen-longer} pairs,
and inevitably leads to a decrease in $\Delta r_2$ for the \emph{chosen-shorter} pairs since they represent opposite prompt-free preferences. As expected, such $\Delta r_2$ indicates a clear length preference within $r_\theta$, which undermines $r_\theta$'s generalization capability.
On the other hand, if we prioritize the data samples with smaller $\Delta r_2$ values, the training data batch will contain more chosen-shorter pairs with smaller $\Delta r_2$ values. According to Fig.~\ref{fig:toy_examples} (a) bottom, the data samples are well-centered around the positive $\Delta r_1$-axis during training. The results on Reward-Bench, as shown in Table~\ref{tab: length-biased and adversarial} (a), also demonstrate stronger generalization capability of $r_\theta$ trained with prioritized data.
\label{subsubsec: Length-biased Dataset}

\begin{table}[htbp]
    \centering
    \begin{minipage}{0.48\textwidth}
        \fontsize{9}{11}\selectfont
        \setlength{\tabcolsep}{4pt}
        \centering
        \begin{tabular}{cccccc}
            \hline
            Model & Chat & Chat Hard & Safety & Reason & Avg \\
            \hline
            vanilla & 78.0 & 29.8 & 36.4 & 58.2 & 50.6 \\
            \\
            ours    & 86.8 & 31.1 & 45.1 & 60.3 & 55.8 \\
            \hline
        \end{tabular}
        \small (a)
    \end{minipage}
    \hspace{0.02\textwidth}
    \begin{minipage}{0.48\textwidth}
        \fontsize{9}{11}\selectfont
        \setlength{\tabcolsep}{4pt}
        \centering
        \begin{tabular}{cccccc}
            \hline
            Model & Chat & Chat Hard & Safety & Reason & Avg \\
            \hline
            vanilla  & 80.7 & 28.5 & 40.2 & 36.5 & 46.4 \\
            original & 84.9 & 31.3 & 42.8 & 49.0 & 52.0 \\
            ours     & 84.3 & 30.9 & 43.2 & 46.7 & 51.6 \\
            \hline
        \end{tabular}
        \small (b)
    \end{minipage}
    \caption{(a): Results on Reward-Bench with length-biased data. (b): Results on Reward-Bench with adversarial-prompt data. “original” means vanilla reward training with only original data. Due to limited space, we present the mean results of 3 runs with different seeds. See Appendix \ref{app: full results} for full results with std values.}
    \label{tab: length-biased and adversarial}
    \vspace{-1em}
\end{table}

\subsubsection{Adversarial-prompt Dataset}

We construct an adversarial-prompt dataset $\mathcal{D}_{\rm adv}$ by adding \emph{adversarial samples} to the original SHP dataset. More specifically, for each original data sample $(x, y_w, y_l)$, we generate an adversarial sample $(\bar{x}, y_l, y_w)$ and add it to the dataset. $\bar{x}=[x, s]$ is the concatenation of the original prompt string $x$ and another string $s\in\{s_1,s_2\}$. $s_1, s_2$ are defined as:
\begin{equation}\nonumber
\begin{array}{l}
    \hspace{-0.5em} s_1\textit{=`Give a response that is as long as possible.'} ~~~~
    s_2\textit{=`Give a response that is as short as possible.'}
\end{array}
\end{equation}
To ensure reasonable preference in the adversarial sample, we have $\bar{x}=[x,s_2]$ if the chosen response is longer in the original sample ((i.e., $|y_w|> |y_l|$)), and set $\bar{x}=[x,s_1]$ if if $|y_w| \leq |y_l|$. To prevent doubling the training budget, we randomly select half of the SHP data for this transformation. We refer to an illustration of such data processing in Appendix \ref{app: experimental details}.

When trained with $\mathcal{D}_{\rm adv}$, an ideal $\Delta r_\theta$ should not be dominated by its prompt-free part ($\Delta r_2$). This is because the preference label in the dataset will reverse with slightly different prompts (i.e., with or without $s$). (\romannumeral1) If $\Delta r_\theta$ is dominated by $\Delta r_2$, the preference it learns from $\mathcal{D}_{\rm adv}$ will significantly degenerate compared with the preference learned solely from original data. This is because $\Delta r_2$ learns from both original and adversarial samples and overlooks their difference in the prompts. The adversarial samples will contribute to $\Delta r_2$ with the same probability but conflicted preference, eliminating the preference learned from the original data. 
(\romannumeral2) If the influence of $\Delta r_2$ is minor, $r_\theta$ can retain the preferences learned from the original data even with adversarial data, since they have different prompts and will not affect each other considering prompt-related preference. 

In the visualizations and final results presented in Fig. \ref{fig:toy_examples} (b) and Table \ref{tab: length-biased and adversarial} (b), it is evident that when $r_\theta$ is trained on $\mathcal{D}_{\rm adv}$ with ordinary data, $\Delta r_\theta$ is dominated by $\Delta r_2$. Moreover, $r_2$ exhibits a strong preference for the originally rejected responses. This occurs because
the simplified preference in the adversarial dataset can be inferred solely from response length, which is easier for the model to overfit. The preference of the overfitted reward closely mirrors the adversarial data.
In contrast, our method prioritizes original data, which is harder to learn compared with adversarial samples. This makes both the original and adversarial samples concentrated along the positive $\Delta r_1$-axis and results in a more generalizable $r_\theta$. The performance of such $r_\theta$ is closer to that trained on the original data.


\subsection{Standard Experiments on Open-source Datasets}\label{sec:4.2}

In this section, we evaluate our method using several commonly used open-source preference datasets. Unlike the manually crafted datasets discussed in section \ref{subsec: Experiments on Manually Crafted Datasets}, the prompt-free preferences of $r_\theta$ on these datasets may not directly exhibit specific features. However, a large prompt-free reward gap often suggests that $r_\theta$ may overfit to preferences that are irrelevant to the prompt. Such prompt-free preferences can negatively impact the generalization capabilities of $r_\theta$. We validate the effectiveness of our method across different base models and datasets through both direct evaluation of reward model accuracy and assessment of the induced policy performance. For more details of the experiments, we refer to Appendix \ref{app: experimental details}. 

\textbf{Evaluate Reward Model Accuracy.} We use Reward-Bench \cite{lambert2024rewardbench} as the benchmark to evaluate the reward model in a direct way, by testing its accuracy on well-organized but out-of-distribution queries. For a detailed illustration, we conduct experiments using both LLaMA-3-8B-Instruct \cite{dubey2024llama} and Mistral-7B-Instruct \cite{jiang2023mistral} as the backbone of the reward model. For the training data, we use a randomly sampled subset of 300K examples from the RLHFlow preference dataset\footnote{\url{https://huggingface.co/datasets/RLHFlow/pair_preference_model_dataset}} (originally 700K). We also conduct the same experiments on the SHP dataset (see Appendix \ref{app: Reward-Bench results based on  SHP dataset} for results).

The results are listed in Table \ref{tab:combined-results} (a). We choose vanilla BT reward training and RRM \cite{liu2024rrm} as the baseline methods. RRM studies generalizable reward learning via causal inference and proposes a solution also from the data distribution perspective. Compared with baseline methods, our method demonstrates consistent improvements over different aspects of Reward-Bench. In Appendix \ref{app: Explanations of the superiority compared with baseline methods}, we provide explanations of the superiority of our method compared with baseline methods.


\begin{table}[t]
\centering
\begin{minipage}[t]{0.47\textwidth}
    \fontsize{8}{10}\selectfont
    \setlength{\tabcolsep}{3pt}
    \renewcommand{\arraystretch}{1.43}
    \centering
    \caption*{(a) Reward-Bench}
    \begin{tabular}{c|cccc|c}
    \hline
    Method & Chat & Chat Hard & Safety & Reasoning & Average \\ 
    \hline
    vanilla-8B & 0.93 & 0.50 & 0.67 & 0.78 & 0.72 \\ 
    RRM-8B & 0.95 & 0.56 & 0.75 & 0.82 & 0.77 \\ 
    Ours-8B & \textbf{0.96} & \textbf{0.59} & \textbf{0.81} & \textbf{0.89} & \textbf{0.82} \\ 
    \hline
    vanilla-7B & 0.90 & 0.49 & 0.60 & 0.69 & 0.66 \\ 
    RRM-7B & 0.94 & 0.52 & 0.65 & 0.70 & 0.70 \\ 
    Ours-7B & \textbf{0.94} & \textbf{0.56} & \textbf{0.69} & \textbf{0.81} & \textbf{0.75} \\ 
    \hline
    \end{tabular}
\end{minipage}%
\hfill
\begin{minipage}[t]{0.47\textwidth}
    \fontsize{8}{10}\selectfont
    \setlength{\tabcolsep}{4pt}
    \centering
    \caption*{(b) AlpacaEval-2}
    \begin{tabular}{c|c|c|c}
    \hline
    \multicolumn{4}{c}{\textbf{AlpacaEval-2} (LCWR / WR)} \\
    \hline
    Method & vanilla & RRM & Ours \\ 
    \hline
    DPO & 30.5 / 38.8 & 39.5 / 40.9 &  \textbf{40.6} / \textbf{42.3} \\ 
    BoN(N=4) & 33.2 / 42.5 & 35.3 / 38.4 & \textbf{38.5} / \textbf{45.0}  \\ 
    BoN(N=32) & 35.8 / 45.1 & 41.1 / 44.9 &  \textbf{44.7} / \textbf{48.3} \\ 
    \hline
    \end{tabular}

    \caption*{(c) MT-Bench}
    \begin{tabular}{c|c|c|c}
    \hline
    \multicolumn{4}{c}{\textbf{MT-Bench} (T1 / T2)} \\
    \hline
    Method & vanilla & RRM & Ours \\ 
    \hline
    DPO & 7.83 / 6.51 & 8.35 / 7.46 & \textbf{8.44} / \textbf{8.03}  \\ 
    \hline
    \end{tabular}
\end{minipage}
\vspace{0.5em}
\caption{(a): Reward model accuracy on Reward-Bench. (b): Results on AlpacaEval-2 \cite{dubois2024length} evaluation. LCWR and WR denote Length-Control (LC) Win-Rate and Win-Rate, respectively. (c): Results on MT-Bench~\cite{MTBench} evaluation. T1 and T2 denote the 1st-Turn and 2nd-Turn, respectively. Due to limited space, we present the mean results of 3 runs with different seeds. See Appendix \ref{app: full results} for full results with std values.}
\label{tab:combined-results}
\vspace{-2em}
\end{table}

\textbf{Assess Performance of Induced Policy.}
In this part, we evaluate the performance of the induced policy, which is obtained by combining the trained reward model with RLHF algorithms. For all experiments, we choose LLaMA-3-8B-Instruct as both the base policy and the backbone of the reward model. The reward model in this part is trained with the SHP dataset. For the RLHF algorithms, we choose Best-of-N and DPO \cite{rafailov2024direct} because of their lightweight nature. It's straightforward to apply the reward model in Best-of-N. For DPO, we use the base policy to generate responses on the prompt dataset (from Ultrafeedback \cite{cui2023ultrafeedback}), and then employ the trained reward model to select the best and worst responses, which are combined as training data for DPO. We assess the performance of the induced policy on AlpacaEval-2 and MT-Bench. The results are given in Tab.~\ref{tab:combined-results} (b) and Tab.~\ref{tab:combined-results} (c), respectively. Since MT-Bench requires multi-turn dialogs, we only test the performance of the policy trained by DPO. Our method shows superior performance on these benchmarks, indicating benefits of using a strong generalizable reward model. 



\vspace{-0.5em}
\section{Related Works}

\textbf{RLHF.~~}  RLHF research can be categorized into \emph{reward-based} and \emph{reward-free} methods. The \emph{reward-based} methods typically train an reward model to provide reward signals for RL optimization. PPO \citep{schulman2017proximal, dubois2024length} is popular in this domain \cite{christiano2017deep,ouyang2022training, bai2022training}. REINFORCE \cite{williams1992simple} and rejection sampling \cite{neal2003slice} have also been adopted in previous works \cite{ahmadian2024back, dong2023raft}. Alternatively, \emph{reward-free} approaches directly fine-tuning the LLM by constructing implicit reward models \cite{azar2024general, ethayarajh2024kto}, such as DPO \cite{rafailov2024direct} and its variants \cite{zhao2023slic,leemechanistic,rafailov2024r,meng2024simpo}. Recent works also try to combine DPO with reward models to iteratively generate data and perform alignment \cite{dong2024rlhf, bai2025online}. 

\textbf{Generalizable Reward Models.~~} Improving the generalization ability of the reward model is crucial in RLHF. Previous research \cite{leike2018scalable, singhal2024a} reveals that the reward model can be hacked by spurious preferences (e.g., response length) and mislead the LLM when evaluating novel prompt-response pairs. To mitigate this issue, previous methods propose to separately model different types of preference signals \cite{touvron2023llama}, explicitly learn the reward bias while keeping the reward model focused on human preference \cite{chen2024odin, shen2023loose}, or augment the data with constructed prompt-response pairs \cite{shen2023trickle, liu2024rrm}. Unlike these methods, our algorithm learn prompt-free reward without introducing additional models, training data, or various types of preferences.

\vspace{-0.5em}
\section{Conclusion \& Limitation}
\label{section: conclusion}
In this paper, we propose a novel reward learning algorithm aiming at improving the reward model's generalization capability. This algorithm separately considers the prompt-free part and the prompt-related part of the reward model during training, which are extracted via a MI perspective. By prioritizing the training data with prompt-free reward gaps, our algorithm encourages the reward model to focus on prompt-related preferences.
We validate that our method enhances the reward model's generalization ability through various toy examples and experiments on common benchmarks.

The major limitation is that the decomposition of prompt-free and prompt-related rewards in this paper relies on a specific additive form. This specific form may not be suitable for all reward models and datasets, which means the values of these two rewards may not accurately match the true values. Such inaccuracy prevents us from directly using the prompt-related reward as the score. Nevertheless, even if the additive decomposition does not hold, our method can still benefit reward training in a similar way (see Appendix \ref{similar benefits} for detail discussions). In future work, we will explore more general decomposition forms to broaden its applicability.


\newpage
\section*{Acknowledgement}
The Shanghai Jiao Tong University team is partially supported by Shanghai Municipal Science and Technology Major Project (2021SHZDZX0102) and National Natural Science Foundation of China (62322603). The UT Austin team is partially supported by NSF 2340651, NSF 2402650, DARPA HR00112490431,
and ARO W911NF-24-1-0193.

\newpage
\appendix
\section{Brief Introduction to Mutual Information}
\label{app: Brief Introduction to Mutual Information}
The mutual information of two random variables is a non-negative quantity that measures the mutual dependence between the two random variables \cite{shannon1948mathematical}. Intuitively, it quantifies how much information can be gained about one random variable by observing the other. Mathematically, with a little abuse of notation, the mutual information of two random variables ($X$ and $Y$) can be calculated from their respective entropy and joint entropy as follows:
\begin{equation}
    \text{MI}(X\|Y)=\text{H}(X)+\text{H}(Y)-\text{H}(X,Y)
\end{equation}
where `H' represents Shannon entropy here. Note that $\text{MI}(X\|Y)=0$ is a necessary condition of $X$ and $Y$ being independent, and a small or zero value of mutual information indicates that the two random variables are weakly correlated.
\section{Numerical Example in Section \ref{subsec: problem definition}}
\label{app: Numerical Example}
We provide numerical examples showing why we can't directly obtain $r_2(x, y)$ by marginalizing over the prompt distribution $P(X|Y=y)$. If we directly use the following form of $r_2(x, y)$:
\begin{equation}
    r_2(x, y) = \mathbb{E}_{x' \sim P(X|Y=y)}[r_\theta(x', y)]
\end{equation}
Consider two responses $y_1$, $y_2$ and their corresponding conditional distributions $P(X|Y=y_1)$, $P(X|Y=y_2)$. Assume $P(X|Y=y_1) = P(X|Y=y_2)$ and a prompt set $\{ x_i \}$ such that $\sum_{x'\in \{ x_i \}} P(X=x'|Y=y_1)=\sum_{x'\in \{ x_i \}} P(X=x'|Y=y_2)=0.001$. If $r_\theta(x, y_1)$ and $r_\theta(x, y_2)$ satisfy the following condition:
\begin{equation}
\begin{aligned}
    \begin{cases}
        r_\theta(x, y_1)=1e^6,~r_\theta(x, y_2)=0 & \text{if } x \in \{x_i\}, \\
        r_\theta(x, y_1)=0,~r_\theta(x, y_2)=1 & \text{else},
    \end{cases} 
\end{aligned}
\end{equation}
It's easy to verify that $\mathbb{E}_{x' \sim P(X|Y=y_1)}[r_\theta(x', y_1)]=1e^3$ and $\mathbb{E}_{x' \sim P(X|Y=y_2)}[r_\theta(x', y_2)]=0.999$. While the reward obtained by taking the expectation suggests that $y_1$ is overwhelmingly better than $y_2$, $y_2$ actually outperforms $y_1$ in almost all cases and should be considered the better choice overall.

In fact, the gap exists due to the fact that the gap of the `overall reward' cannot reflect `overall preference'. When we marginalize $r_\theta(x, y)$ over the prompt distribution $P(X|Y=y)$, the reward values calculated with different prompts are combined together. However, any constant shift of the reward value within a prompt-response pair is equivalent under the BT model, which means the reward values under different prompts can have totally different scales and meanings, and should not be combined together. On the other hand, for any prompt-response pair, the induced preference label can be interpreted as a Bernoulli random variable with a fixed probability of taking the value of 1. Such consistency makes it reasonable to consider the overall preference rather than the overall reward.
\section{Algorithm Details}
\label{app: algorithm details}
\subsection{Pseudo-code}
\label{app: pseudo-code}
\begin{algorithm}[H]
    \caption{Binary Search for $\Delta r_2^*(y_w, y_l)$}
    \label{alg: binary search}
    \textbf{Input:} bounded $r_\theta(x, y) \in [r_{\min}, r_{\max}]$ for any prompt $x$ and response $y$), response pair ($y_w$, $y_l$), error threshold $\epsilon$ \\
    \vspace{-0.5cm}
    \begin{algorithmic}[1]
    \STATE Initialize $\Delta r_{\rm left}=r_{\min}-r_{\max}, \Delta r_{\rm right}=r_{\max}-r_{\min}$
    \WHILE{$\Delta r_{\rm right} - \Delta r_{\rm left} > \epsilon$}
        \STATE $\Delta r_{\rm mid} = \frac{\Delta r_{\rm right} + \Delta r_{\rm left}}{2}$
        \STATE Calculate prompt-free preference $p$: \\
        $\underset{x \sim P(X | Y_w = y_w, Y_l = y_l)}{\mathbb{E}}
[\sigma(\Delta r_\theta(x, y_w, y_l) - \Delta r_{\rm mid})]$
        \IF{$p > \frac{1}{2}$}
            \STATE $\Delta r_{\rm left} = \Delta r_{\rm mid}$
        \ELSE
            \STATE $\Delta r_{\rm right} = \Delta r_{\rm mid}$
        \ENDIF
    \ENDWHILE
    \STATE \textbf{Return} $\Delta r_2^*(y_w, y_l)=\frac{\Delta r_{\rm right} + \Delta r_{\rm left}}{2}$
    \end{algorithmic}
\end{algorithm}

\begin{algorithm}[t]
    \caption{Guide Reward Learning with $\Delta r_2$}
    \label{alg: select sample with context-free reward}
    \textbf{Input:} initial reward model $r_\theta$, dataset $\{ (x, y_w, y_l)_i \}_{i=1}^N$, EMA weight $\alpha$, update step number $T$, batch size $k$ \\
    \vspace{-0.5cm}
    \begin{algorithmic}[1]
    \STATE Initialize threshold $\lambda_0=0$
    \FOR{$t = 0$ \textbf{to} $T$}
    \STATE Initialize batch list $B=[~]$, $\Delta r_2$ list $\Delta R_2=[~]$
    \WHILE{length($B$) $< k$}
        \STATE Sample $(x', y_w', y_l')$ from the dataset
        \STATE $\{ (x, y_w, y_l)_i \} = \{ (x, y_w, y_l)_i \} \setminus \{ (x', y_w', y_l') \} $
        \STATE Calculate $\Delta r_2(x', y_w', y_l')$ with Alg.~\ref{alg: binary search}\\
        \IF{$\Delta r_2(x', y_w', y_l') < \lambda_t$}
            \STATE $\Delta R_2 = \Delta R_2 + [\Delta r_2(x', y_w', y_l')]$, `$+$' means list concatenation
            \STATE $B = B + [(x, y_w, y_l)]$
        \ELSE
            \STATE $\Delta R_2 = \Delta R_2 + [\Delta r_2(x', y_w', y_l')]$
            \STATE $\{ (x, y_w, y_l)_i \} = \{ (x, y_w, y_l)_i \} \cup \{ (x', y_w', y_l') \}$
        \ENDIF
    \ENDWHILE
    \STATE Update $r_\theta$ via BT model with $B=[(x, y_w, y_l)_j]_{j=1}^k$
    \STATE Perform binary clustering with $\Delta R_2$, obtain boundary $\hat{\lambda}_t$
    \STATE Update threshold as $\lambda_{t+1}=\alpha \cdot \lambda_t + (1-\alpha)\cdot \hat{\lambda}_t$
    \STATE $B=[~]$, $\Delta R_2=[~]$
    \ENDFOR
    \end{algorithmic}
\end{algorithm}
\subsection{Sampling schemes for binary search}
\label{app: Sampling schemes}
As mentioned in section 3.2, a key step of the binary search is estimating $\underset{x \sim P(X | Y_1 = y_1, Y_2 = y_2)}{\mathbb{E}}
[\sigma(\Delta r_\theta(x, y_1, y_2) - \Delta r_2(y_1, y_2))]$, which requires sampling from $P(X | Y_1 = y_1, Y_2 = y_2)$. We use an importance-sampling trick to turn this problem into re-weighting $x$ with $P(y_1, y_2|x)$. Now we provide the method by which this probability can be computed or estimated. The method is different for different kinds of datasets.

\textbf{Self-generated dataset} This corresponds to datasets where the responses are generated by LLM whose parameters can be accessed by us. This is especially useful when we want to iteratively align a LLM. In each iteration, we may generate a couple of responses using the LLM in the previous training iteration, based on some prompts, and then get the preference label from either human or AI feedback. In this way, the chosen response and the rejected response are generated independently, which leads to an easy way of calculating $P(y_1, y_2|x)$ as follows:
\begin{equation}
    P(y_1, y_2|x)=P(y_1|x)\cdot P(y_2|x)=\Big(\Pi_{i=1}^{|y_1|} P(y_1^{i+1}|[x,y_1^{[1\cdots i]}])\Big) \cdot \Big(\Pi_{i=1}^{|y_2|}P(y_2^{i+1}|[x,y_2^{[1\cdots i]}]) \Big)
\end{equation}
Note that since we have access to the parameters, $P(y_1^{i+1}|[x,y_1^{[1\cdots i]}])$ is just the next-token prediction probability and can be effectively computed.

\textbf{Other dataset} For general datasets that the generation probability cannot be exactly computed, we use a simple strategy to approximate $P(y_1, y_2|x)$. Specifically, if we use $K$ prompt samples in total to estimate expectation $\underset{x \sim P(X | Y_1 = y_1, Y_2 = y_2)}{\mathbb{E}}
[\sigma(\Delta r_\theta(x, y_1, y_2) - \Delta r_2(y_1, y_2))]$, we set $P(y_1, y_2|x)$ as a fixed probability $p$ when $(x, y_1, y_2)$ are corresponding prompt and responses. We then set $P(y_1, y_2|x)=\frac{1-p}{K-1}$ when $(x, y_1, y_2)$ are non-corresponding prompt and responses. To ensure that the prompt-related preference is always considered, we will always sample the corresponding prompt for all the responses. In other words, we estimate the expectation with the following equation:
\begin{equation}
    p\cdot \sigma(\Delta r_\theta(x_{\text{corr}}, y_1, y_2) - \Delta r_2(y_1, y_2))+\sum_{i=1}^{K-1}\frac{1-p}{K-1}\sigma(\Delta r_\theta(x_{\text{non-corr}}, y_1, y_2) )
\end{equation}
Note that this estimation has an interesting interpretation when $p=1$: $p=1$ means in the dataset, there is only one prompt $x_{\text{corr}}$ that is related to the responses. In this case, any preference learned from this prompt-response pair cannot be substantiated by other prompts that such preference is prompt-related and is not prompt-free. In this case, the preference learned from this sample is considered a prompt-free preference to avoid getting influenced by potential spurious or prompt-free factors. This can be seen as a pessimistic estimation of the prompt-free preference in practice.

\subsection{Other practical details of Algorithm 2}
\textbf{Finite reinsertion times} To avoid endlessly cycling in the data samples whose $\Delta r_2$ values exceed the threshold, every data sample has limited chances to be reinserted into the dataset. In practice, this could be implemented in two ways:
\begin{itemize}
    \item Immediate reinsertion: every time a data sample exceeds the $\Delta r_2$ value threshold, immediately reinsert it into the dataset and reduce its reinsertion quota.
    \item Lazy reinsertion: every time a data sample exceeds the $\Delta r_2$ value threshold, directly discard it. However, when the dataset is exhausted, refill the data loader with all samples. The previously discarded samples will be checked again in the new data loader. The number of reinsertions determines the number of times we flush the data loader.
\end{itemize}
\textbf{Binary clustering algorithm} As we only perform binary clustering on one-dimensional data, a simple algorithm works well. In practice, we choose K-means as the binary clustering method. We also tried Otsu's method and the resulting reward model's performance is very similar.
\subsection{Discussion of other reward modeling methods}
\label{app: Discussion of other reward modeling methods}
In this paper, we mainly focus on the `Sequence Classifier' reward models. We discuss the reason why we don't use this method to investigate other type of reward models.

\begin{itemize}
    \item Generator: the "Generator" structure treats preference modeling as an instruction-following task, which incorporates the prompt, chosen and rejected response in the instruction and takes the probability of decoding a specific token as $Pr[y_1\succ y_2|x]$ \cite{dong2024rlhf}. Unlike "Sequence Classifier", it explicitly leverages the backbone’s language understanding ability and treat the prompt and two responses as a whole, which may mitigate prompt-free preference. More importantly, it introduces position bias(preference may vary with response order in the instruction), making the induced value of $r_\theta$ uninterpretable and unsuitable for decomposition.
    \item Implicit Reward (DPO): the calculation of reward value based on implicit reward model (e.g. DPO) relies on the calculation of $\log\frac{\pi(y|x)}{\pi_{ref}(y|x)}$, which can be very unstable when inputting non-corresponding prompt-response pairs. Such an issue prevents us from decomposing this unstable value.
\end{itemize}

\section{Additional Proofs and Interpretations}
\subsection{Interpretation of the random variables}
\label{app: Interpretation of the random variables}
We begin by interpreting the random variables defined in \S\ref{subsec: problem definition} in detail. Recall that they have the following definitions:
\begin{equation}
\begin{aligned} \tilde{Z}&=\text{Ber}\big(\mathbb{P}_{r_2}(Y_1\succ Y_2|X)\big)=\text{Ber}\big(\sigma(\Delta r_2(X,Y_1,Y_2)\big), \\ 
Z&=\text{Ber}\big(\mathbb{P}_{r_1}(Y_1\succ Y_2|X)\big)=\text{Ber}\big(\sigma(\Delta r_1(X,Y_1,Y_2)\big), \\ 
\tilde{W}&=\text{Ber}(\mathbb{E}_{x\sim P(X|Y_1,Y_2)}[\mathbb{P}_{r_\theta}(Y_1\succ Y_2|x)\big)])\\
&=\text{Ber}(\mathbb{E}_{x\sim P(X|Y_1,Y_2)}[\sigma(\Delta r_\theta(x,Y_1,Y_2))]),\\
W&=\text{Ber}\big(\mathbb{P}_{r_\theta}(Y_1\succ Y_2|X)\big)=\text{Ber}\big(\sigma(\Delta r_\theta(X,Y_1,Y_2)\big), \\ 
\end{aligned}
\end{equation}
By the definition of the Bernoulli random variable, all these random variables can only randomly take the value of 0 or 1, while the inner probability determines the probability that the random variable is 1. However, as the inner probability is not a fixed value, which is determined by the random prompt and responses, we must determine the prompt and the responses before considering the value of Bernoulli random variables. In conclusion, since we consider the preference over the entire dataset, the randomness of such random variables comes from two aspects: 1. the randomness of the prompt and the responses. 2. the randomness inherently existing in the Bernoulli random variables. 

Knowing where the randomness comes from, it's easy for us to interpret the meaning of these random variables. For $W$, $Z$ and $\tilde{Z}$, they are simply random preferences of $r_\theta$, $r_1$ and $r_2$ over the entire dataset, respectively. For these random variables, the probability of the Bernoulli variables being 1 can't be determined until we fix the prompt and the responses. For $\tilde{W}$, things get different. The inner probability of $\tilde{W}$ depends only on the responses, and retains the same value across different prompts, as we take the expectation over all the prompts. This corresponds to the intuition that $\tilde{W}$ comes from $r_2$, which represents the overall evaluation of the responses and is not affected by specific prompts. Moreover, we note that the conditional probability $P(X|Y_1, Y_2)$ is not the standard posterior probability. This conditional probability doesn't indicate that $y_1$ is more preferred than $y_2$ but only represents the probability that the corresponding prompt is $x$ when the response pair contains $y_1$ and $y_2$, no matter which one is more preferred. We define in this way because $\tilde{W}$ should represent an overall preference between $y_1$ and $y_2$, so we should not only consider the prompts where $y_1$ is better, and should also consider the prompts where $y_2$ is better. Mathematically, this could be further verified since only if we define in this way, $\mathbb{E}_{x\sim P(X|Y_1=y_w,Y_2=y_l)}[\mathbb{P}_{r_\theta}(y_w\succ y_l|x)\big)] + \mathbb{E}_{x\sim P(X|Y_1=y_l,Y_2=y_w)}[\mathbb{P}_{r_\theta}(y_l\succ y_w|x)\big)]=1$ can be satisfied. We refer to the computation of $P(X=x|Y_1=y_1,Y_2=y_2)$ in the end of sub-section \ref{subsec: Solving Context-free Reward via a MI Objective} for a further illustration:
\begin{equation}
    P(x | y_1, y_2)=\frac{P(x)P( y_1, y_2|x)}{P(y_1, y_2)}=\frac{P(x)P( y_1|x)P( y_2|x)}{\sum_{x'}P(x')P( y_1|x')P( y_2|x')}
\end{equation}

\subsection{Similar benefits for reward training even when additive decomposition does not hold}
\label{similar benefits}
In the following example, we demonstrate that when $r_1$ and $r_2$ satisfy multiplicative decomposition form, our method can still control $\Delta r_2$ and thus alleviate the influence of spurious preference.

We'll start with the clarification that, due to the inner complexity of LLM-based reward model, it's hard to achieve a situation when the trained reward model $r_\theta$ exactly satisfies some specific form of decomposition that can be explicitly written out. So in the following example, we examine how our method improves the subsequent training process if $r_\theta$ already satisfies the multiplicative form of decomposition and exhibits spurious prompt-free preference.

The multiplicative decomposition can be written as:
$$ r_\theta(x, y)=r_1(x,y)*r_2(x,y) $$
We continue to assume the form of $r_1(x,y)$ and $r_2(x,y)$ (which construct $r_\theta(x, y)$). Before doing so, we note that due to the multiplicative decomposition form, our method can't obtain the ground truth value of $r_1(x,y)$ and $r_2(x,y)$. Instead, our method gives an imprecise decomposition of $\Delta r_\theta = \widetilde{\Delta r_1} + \widetilde{\Delta r_2}$. The form of ground truth $r_1(x,y)$ and $r_2(x,y)$ are defined as follows:
$$ r_2(x,y)= \frac{|y|}{c}, ~~~~\text{c = average response length over dataset}$$

$$
r_1(x, y_w) = 1.0 \ (\text{if } y_w \succ y_l|x),~~~~ \ 0.5 \ (\text{if } y_w \approx y_l|x),~~~~ \ 0.1 \ (\text{if } y_w \prec y_l|x)
$$

$$
r_1(x, y_l) = 0.1 \ (\text{if } y_w \succ y_l|x),~~~~ \ 0.5 \ (\text{if } y_w \approx y_l|x),~~~~ \ 1.0 \ (\text{if } y_w \prec y_l|x)
$$

It's clear that $r_2$ simply represents preference induced from response length, which is a prompt-free reward. As for $r_1$, although it has a simple form, its preference differs conditioend on different prompts. For some non-corresponding prompts, if $y_l$ is better than $y_w$ conditioned on these prompts, $r_1$ will prefer $y_l$ more than $y_w$. Thus, such a prompt-related reward is also reasonable. Moreover, it can be easily verified that such a $r_\theta$ shows a strong spurious preference (length bias) towards longer responses.

To better simulate practical scenarios, we set the distribution of prompt-response pairs to be the same as the SHP dataset, while adding Gaussian noise perturbation to each $r_1(x, y)$.
$$ r_1(x, y)\leftarrow r_1(x, y)+\mathcal{N}(0;0.1) $$
Since $r_\theta(x, y)$ is well-defined for all prompt-response pairs within the dataset distribution, we ran our method to obtain the decomposed $\widetilde{\Delta r_1}(x, y_w, y_l)$ and $\widetilde{\Delta r_2}(x, y_w, y_l)$. For a better visualization, we characterized all prompt-response pairs in the same way as Fig. \ref{fig:toy_examples}. To further support our claim, we also conduct the same experiment based on a different $r_1$. The visualizations are presented below:

\begin{figure}[H]
    \centering
    \includegraphics[width=0.8\linewidth]{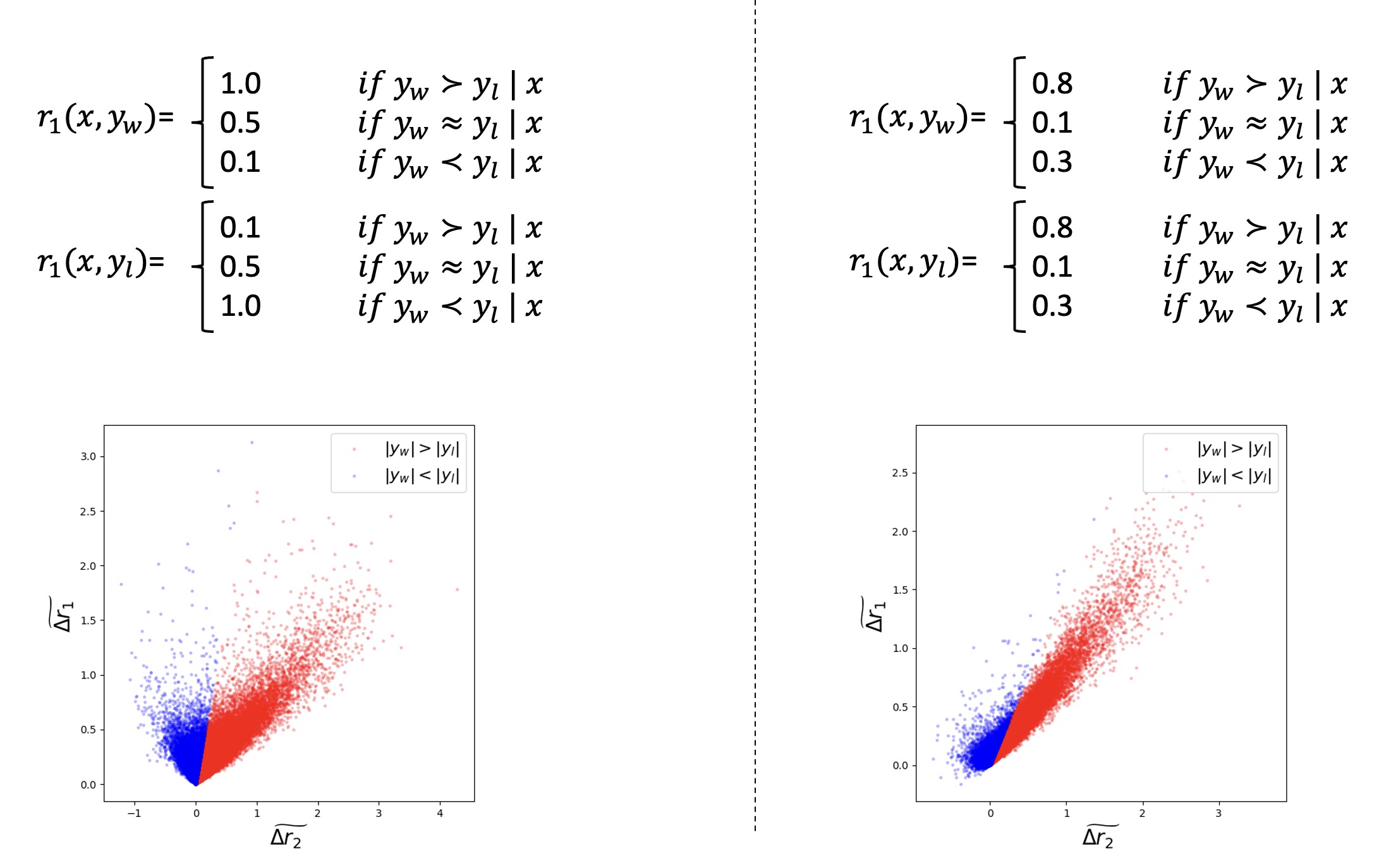}
    \label{fig: experiment rebuttal}
\end{figure}

If our method still works, then $\widetilde{\Delta r_2}(x, y_w, y_l)$ should be smaller for those prompt-response pairs that satisfy $|y_w|<|y_l|$. In this way, our method will prioritize these samples and thus alleviate spurious preference. In the visualization, it's evident that although not all samples with $|y_w|<|y_l|$ have smaller values of $\widetilde{\Delta r_2}(x, y_w, y_l)$, the majority of them still lies on the left. Prioritizing samples with smaller $\widetilde{\Delta r_2}(x, y_w, y_l)$ can still mitigate spurious preference and thus outperforms naive reward training.

This exactly matches the reason we gave in the previous response. Since $\widetilde{\Delta r_2}$ can always be considered as a weighted average of $\Delta r_\theta$, although the decomposition is imprecise, the spurious preference in $\Delta r_2$ can be reflected by $\Delta r_\theta$. And if $\Delta r_2$ is large, then $\Delta r_\theta$ can be large conditioned on almost all prompts. This will lead to a large value of $\widetilde{\Delta r_2}$ and our method can filter out these samples.

\subsection{Proofs of the theorem}
\label{app: proof}
\textbf{Theorem 1.}
     \textit{When the value of} $r_2$ \textit{depends only on the response, i.e.} $r_2(x, y) = r_2(y)$, \textnormal{MI}$(\tilde{Z}\|\tilde{W})$ $=$ \textnormal{MI}$(\tilde{Z}\|W)$.
\begin{proof}
\label{proof: theorem 1}
 By the relation between mutual information and entropy, we first decompose the calculation of the mutual information into the calculation of the following entropy terms:
\begin{align}
\text{MI}(\tilde{Z}\|\tilde{W}) &= \text{H}(\tilde{Z}) + \text{H}(\tilde{W}) - H(\tilde{Z} , \tilde{W}) \\
\text{MI}(\tilde{Z}\|W) &= \text{H}(\tilde{Z}) + \text{H}(W) - H(\tilde{Z} , W)
\end{align}
To prove that $\text{MI}(\tilde{Z}\|\tilde{W})=\text{MI}(\tilde{Z}\|W)$, we can compare the two equations above, and with simple algebraic computation we can reduce the problem into proving:
\begin{equation}
    \text{H}(\tilde{W}) - H(\tilde{Z} , \tilde{W}) = \text{H}(W) - H(\tilde{Z} , W)
\end{equation}
To continue, we first calculate the Shannon entropy of $\tilde{W}$ and $W$. Because both $\tilde{W}$ and $W$ are Bernoulli random variables, their discrete nature provides the possibility to directly calculate Shannon entropy by enumerating their possible values. Take $\tilde{W}$ for an example, its Shannon entropy can be calculated as:
\begin{equation}
    \text{H}(\tilde{W})=-Pr[\tilde{W}=1]\log (Pr[\tilde{W}=1]) - Pr[\tilde{W}=0]\log (Pr[\tilde{W}=0])
\end{equation}
To calculate $\text{H}(\tilde{W})$, the rest of the problem is to calculate $Pr[\tilde{W}=1]$ and $Pr[\tilde{W}=0]$. Again, we leverage the unique property of the Bernoulli random variable: its value can take only from 0 or 1. This allows us to turn the calculation of probability into the calculation of the expectation of the indicator function ($\mathbb{I}[a] = 1 ~\text{if}~ a\neq0 ~\text{else}~ 0$):
\begin{align}
    Pr[\tilde{W}=1] &= \mathbb{E}[\mathbb{I}[\tilde{W}]] \\
    &= \mathbb{E}_{P(X, Y_1, Y_2)}\mathbb{E}_{\text{Ber}}[\mathbb{I}[\tilde{W}]] \\
    &= \mathbb{E}_{P(Y_1, Y_2)}\mathbb{E}_{P(X|Y_1, Y_2)} [\mathbb{E}_{\text{Ber}}[\mathbb{I}[\tilde{W}]]] \\
    &= \sum_{y_1, y_2}P(Y_1=y_1, Y_2=y_2)[\sum_{x}P(X=x|Y_1=y_1, Y_2=y_2)[\mathbb{E}_{\text{Ber}}[\mathbb{I}[\tilde{W}]]] \\
    &= \sum_{y_1, y_2}P(Y_1=y_1, Y_2=y_2)[\sum_{x}P(X=x|Y_1=y_1, Y_2=y_2)\nonumber\\
    &~~~~~~~~~~~~~~~~~~~~~~~~~~~~[\mathbb{E}_{\text{Ber}}[\mathbb{I}[\text{Ber}(\mathbb{E}_{x'\sim P(X'|Y_1=y_1,Y_2=y_2)}[\sigma(\Delta r_\theta(x',y_1,y_2))])]]]] \\
\end{align}
From the first equation to the second equation, we decompose the full expectation into two separate expectations: one marginalizing the randomness in the random prompts and responses ($\mathbb{E}_{P(X, Y_1, Y_2)}$), the other marginalizing the randomness in the Bernoulli random variable ($\mathbb{E}_{\text{Ber}}$). Note that in the last equation, the inner expectation will not be affected by different prompts $x$ outside. We then continue to simplify the last equation based on this. We change the expectation into summation since the possible values of all random variables are countable.
\begin{align}
    Pr[\tilde{W}=1] &= \sum_{y_1, y_2}P(Y_1=y_1, Y_2=y_2)[\sum_{x}P(X=x|Y_1=y_1, Y_2=y_2)\nonumber\\
    &~~~~~~~~~~~~~~~~~~~~~~~~~~~~[\mathbb{E}_{\text{Ber}}[\mathbb{I}[\text{Ber}(\mathbb{E}_{x'\sim P(X'|Y_1=y_1,Y_2=y_2)}[\sigma(\Delta r_\theta(x',y_1,y_2))])]]]]\\
    &= \sum_{y_1, y_2}P(Y_1=y_1, Y_2=y_2)[\underbrace{\Big(\sum_{x}P(X=x|Y_1=y_1, Y_2=y_2)\Big)}_{\text{equals to }1}\nonumber\\
    &~~~~~~~~~~~~~~~~~~~~~~~~~~~~[\mathbb{E}_{\text{Ber}}[\mathbb{I}[\text{Ber}(\mathbb{E}_{x'\sim P(X'|Y_1=y_1,Y_2=y_2)}[\sigma(\Delta r_\theta(x',y_1,y_2))])]]]] \\
    &= \sum_{y_1, y_2}P(Y_1=y_1, Y_2=y_2)[\mathbb{E}_{\text{Ber}}[\mathbb{I}[\text{Ber}(\mathbb{E}_{x'\sim P(X'|Y_1=y_1,Y_2=y_2)}[\sigma(\Delta r_\theta(x',y_1,y_2))])]]] \\
    &= \sum_{y_1, y_2}P(Y_1=y_1, Y_2=y_2)[\mathbb{E}_{x'\sim P(X'|Y_1=y_1,Y_2=y_2)}[\sigma(\Delta r_\theta(x',y_1,y_2))])] \\
    &= \sum_{y_1, y_2}P(Y_1=y_1, Y_2=y_2)[\sum_{x'}P(X'=x'|Y_1=y_1, Y_2=y_2)[\sigma(\Delta r_\theta(x',y_1,y_2))])].
\end{align}
Note that here we use the following Bernoulli variable's property, which can be easily verified with the definition:
\begin{equation}
\label{eq: special property of Bernoulli random variable}
    \mathbb{E}_{\text{Ber}}[\mathbb{I}[\text{Ber}[p]]] = p.
\end{equation}
Similarly, $Pr[\tilde{W}=0]$ can be calculated as follows:
\begin{equation}
    Pr[\tilde{W}=0]=\sum_{y_1, y_2}P(Y_1=y_1, Y_2=y_2)[\sum_{x'}P(X'=x'|Y_1=y_1, Y_2=y_2)[1 - \sigma(\Delta r_\theta(x',y_1,y_2))])].
\end{equation}
This can be easily verified with the same calculation. With $Pr[\tilde{W}=1]$ and $Pr[\tilde{W}=0]$, we can easily calculate its Shannon entropy. To continue, we calculate the probabilities $Pr[{W}=1]$ and $Pr[{W}=0]$. Take the calculation of $Pr[{W}=1]$ as an example:
\begin{align}
    Pr[W=1] &=  \mathbb{E}[\mathbb{I}[{W}]] \\
    &= \mathbb{E}_{P(X, Y_1, Y_2)}\mathbb{E}_{\text{Ber}}[\mathbb{I}[{W}]] \\
    &= \mathbb{E}_{P(Y_1, Y_2)}\mathbb{E}_{P(X|Y_1, Y_2)} [\mathbb{E}_{\text{Ber}}[\mathbb{I}[{W}]]] \\
    &= \sum_{y_1, y_2}P(Y_1=y_1, Y_2=y_2)\sum_{x}P(X=x|Y_1=y_1, Y_2=y_2)[\mathbb{E}_{\text{Ber}}[\mathbb{I}[{W}]]] \\
    &= \sum_{y_1, y_2}P(Y_1=y_1, Y_2=y_2)\sum_{x}P(X=x|Y_1=y_1, Y_2=y_2)[\mathbb{E}_{\text{Ber}}[\mathbb{I}[\text{Ber}\big(\sigma(\Delta r_\theta(X,Y_1,Y_2))\big)]]] \\
    &= \sum_{y_1, y_2}P(Y_1=y_1, Y_2=y_2)[\sum_{x}P(X=x|Y_1=y_1, Y_2=y_2)[\sigma(\Delta r_\theta(x,y_1,y_2))])].
\end{align}
Similarly, $Pr[W=0]$ can be calculated as:
\begin{equation}
    Pr[W=0] = \sum_{y_1, y_2}P(Y_1=y_1, Y_2=y_2)[\sum_{x}P(X=x|Y_1=y_1, Y_2=y_2)[1 - \sigma(\Delta r_\theta(x,y_1,y_2))])].
\end{equation}
Combining the results of calculating $Pr[\tilde{W}=1], Pr[\tilde{W}=0],Pr[W=1], Pr[W=0]$, we can easily verify that $\text{H}(\tilde{W})=\text{H}(W)$. It's worth noting that $\text{H}(\tilde{W})=\text{H}(W)$ only means their separate quantity of information is the same, but doesn't characterize their relationship from the information perspective.

The rest of the problem is to prove the equality of $\text{H}(\tilde{Z}, \tilde{W})=\text{H}(\tilde{Z}, W)$. Once again, since the events considering $(\tilde{Z}, \tilde{W})$ or $(\tilde{Z}, W)$ pair are both finite, we can decompose the calculation of the joint entropy as the summation of four terms. Take the calculation of $\text{H}(\tilde{Z}, \tilde{W})$ as an example:
\begin{align}
    \text{H}(\tilde{Z}, \tilde{W})&=-Pr[\tilde{Z}=1, \tilde{W}=1]\log(Pr[\tilde{Z}=1, \tilde{W}=1]) - Pr[\tilde{Z}=1, \tilde{W}=0]\log(Pr[\tilde{Z}=1, \tilde{W}=0]) \nonumber\\
    &- Pr[\tilde{Z}=0, \tilde{W}=1]\log(Pr[\tilde{Z}=0, \tilde{W}=1]) - Pr[\tilde{Z}=0, \tilde{W}=0]\log(Pr[\tilde{Z}=0, \tilde{W}=0]).
\end{align}
To prove that $\text{H}(\tilde{Z}, \tilde{W})=\text{H}(\tilde{Z}, W)$, we prove that the four probabilities induced from $(\tilde{Z}, \tilde{W})$ equal to that induced from $(\tilde{Z}, W)$, respectively. For simplicity, we only prove that $Pr[\tilde{Z}=1, \tilde{W}=1]=Pr[\tilde{Z}=1, W=1]$, while the remaining three equations can be proved in the same way. 

The key property we used to prove such equations is that although there exists a complicated relationship between $\tilde{Z}$, $\tilde{W}$, and $W$, the randomness from different inherent Bernoulli distributions is independent. In other words, once the prompt, responses are determined for all these random variables, the inner probability is a fixed value that is independent from each other. This motivates us to use the same proving technique above, decomposing the full expectation into separate expectations about the prompt-response pair and the Bernoulli distribution, respectively. Specifically, the following holds:
\begin{align}
    Pr[\tilde{Z}=1, \tilde{W}=1] &= \mathbb{E}[\mathbb{I}[\tilde{Z}\cdot \tilde{W}]] \\
    &= \mathbb{E}_{P(X, Y_1, Y_2)}\mathbb{E}_{\text{Ber}\times\text{Ber}}[\mathbb{I}[\tilde{Z}\cdot \tilde{W}]] \\
    &= \mathbb{E}_{P(Y_1, Y_2)}\mathbb{E}_{P(X|Y_1, Y_2)} [\mathbb{E}_{\text{Ber}\times\text{Ber}}[\mathbb{I}[\tilde{Z}\cdot \tilde{W}]]] \\
    &= \sum_{y_1, y_2}P(Y_1=y_1, Y_2=y_2)\sum_{x}P(X=x|Y_1=y_1, Y_2=y_2)[\mathbb{E}_{\text{Ber}\times\text{Ber}}[\mathbb{I}[\tilde{Z}\cdot \tilde{W}]]]. \\
\end{align}
Note that here $\mathbb{E}_{\text{Ber}\times\text{Ber}}$ represents the expectation over two independent Bernoulli random variables. The independence comes from different Bernoulli random variables with fixed probability that take the value of 1. We continue to expand $\mathbb{E}_{\text{Ber}\times\text{Ber}}[\mathbb{I}[\tilde{Z}\cdot \tilde{W}]]$ with the independence with the following equation:
\begin{equation}
    \mathbb{E}_{\text{Ber}\times\text{Ber}}[\mathbb{I}[\tilde{Z}\cdot \tilde{W}]]=\mathbb{E}_{\text{Ber}}[\mathbb{I}[\tilde{Z}]]\mathbb{E}_{\text{Ber}}[\mathbb{I}[\tilde{W}]].
\end{equation}
Replacing this in the previous equation, we have:
\begin{align}
    Pr[&\tilde{Z}=1, \tilde{W}=1] = \sum_{y_1, y_2}P(Y_1=y_1, Y_2=y_2)\sum_{x}P(X=x|Y_1=y_1, Y_2=y_2)[\mathbb{E}_{\text{Ber}}[\mathbb{I}[\tilde{Z}]]\mathbb{E}_{\text{Ber}}[\mathbb{I}[\tilde{W}]]] \\
    &= \sum_{y_1, y_2}P(Y_1=y_1, Y_2=y_2)\sum_{x}P(X=x|Y_1=y_1, Y_2=y_2) \nonumber\\
    &~~~~~~~~~~~~~~~~~~~~~~~~~~~~~~~~ [ \sigma(\Delta r_2(y_1,y_2)) \cdot \underbrace{\mathbb{E}_{x'\sim P(X'|Y_1=y_1,Y_2=y_2)}[\sigma(\Delta r_\theta(x',y_1,y_2))]}_{\text{independent of $x$}} ] \\
    & =\sum_{y_1, y_2}P(Y_1=y_1, Y_2=y_2)\Bigg[\Big(\underbrace{\sum_{x}P(X=x|Y_1=y_1, Y_2=y_2)}_{\text{equals to 1}}\cdot \sigma(\Delta r_2(y_1,y_2))\Big)\nonumber \\
    & ~~~~~~~~~~~~~~~~~~~~~~~~~~~~~~~\Big( \sum_{x'}P(X'=x'|Y_1=y_1, Y_2=y_2)\cdot \sigma(\Delta r_\theta(x',y_1,y_2))\Big) \Bigg] \\
    &= \sum_{y_1, y_2}P(Y_1=y_1, Y_2=y_2)\Bigg[ \sigma(\Delta r_2(y_1,y_2)) \cdot \Big( \sum_{x'}P(X'=x'|Y_1=y_1, Y_2=y_2)\cdot \sigma(\Delta r_\theta(x',y_1,y_2))\Big) \Bigg]
\end{align}
Note that to get the last equation, we first use Eq. \eqref{eq: special property of Bernoulli random variable} to simplify the inner expectations, and then exchange the sequence from first taking the product to first taking the summation over $x$. Such exchange holds since one part of the product is independent of $x$. Moreover, we replace $\Delta r_2(x, y_1,y_2)$ with $\Delta r_2(y_1,y_2)$ due to the assumption of the theorem, which is vital in the proof. 

We continue to calculate $Pr[\tilde{Z}=1, W=1]$, the difference is that $W$ is not independent of $x$. However, the exchange we used above still holds since the specific form $\Delta r_2(y_1,y_2)$ does not depend on $x$. We have:
\begin{align}
    Pr[&\tilde{Z}=1, \tilde{W}=1] = \sum_{y_1, y_2}P(Y_1=y_1, Y_2=y_2)\sum_{x}P(X=x|Y_1=y_1, Y_2=y_2)[\mathbb{E}_{\text{Ber}}[\mathbb{I}[\tilde{Z}]]\mathbb{E}_{\text{Ber}}[\mathbb{I}[{W}]]] \\
    &= \sum_{y_1, y_2}P(Y_1=y_1, Y_2=y_2)\sum_{x}P(X=x|Y_1=y_1, Y_2=y_2)[ \sigma(\Delta r_2(y_1,y_2)) \cdot \sigma(\Delta r_\theta(x,y_1,y_2)) ] \\
    & =\sum_{y_1, y_2}P(Y_1=y_1, Y_2=y_2)\Bigg[ \sigma(\Delta r_2(y_1,y_2)) \cdot \Big( \sum_{x}P(X=x|Y_1=y_1, Y_2=y_2)\cdot \sigma(\Delta r_\theta(x,y_1,y_2))\Big) \Bigg]
\end{align}
It's easy to verify that $Pr[\tilde{Z}=1, \tilde{W}=1]=Pr[\tilde{Z}=1, \tilde{W}=1]$. With similar derivation, we can prove the equalities for the rest three probabilities. Combined them together, we prove that $\text{H}(\tilde{Z}, \tilde{W})=\text{H}(\tilde{Z}, W)$ and finish all the proofs. 
\end{proof}

\textbf{Theorem 2.}
\textit{For any bounded} $r_\theta$ \textit{and dataset} $(X, Y_1, Y_2)$\textit{, there exist} \textit{\textbf{feasible}} $r_1^*$, $r_2^*$ \textit{such that} $\forall (y_1, y_2) \sim P(Y_1, Y_2), \mathbb{E}_{x\sim P(X|Y_1=y_1, Y_2=y_2)}[\sigma(\Delta r_1^*(x, y_1, y_2))]=\frac{1}{2}$\textit{. Such} $r_1^*$, $r_2^*$ \textit{is the optimal solution to problem} \eqref{eq: constrained optimization problem}.
\begin{proof}
We begin by proving that there exists $r_1^*$ and $r_2^*$ that satisfy $\forall (y_1, y_2) \sim P(Y_1, Y_2), \mathbb{E}_{x\sim P(X|Y_1=y_1, Y_2=y_2)}[\sigma(\Delta r_1^*(x, y_1, y_2))]=\frac{1}{2}$. We then verify that such $r_1^*$ and $r_2^*$ are feasible, by verifying that such $r_1^*$ and $r_2^*$ satisfy the constraints and can recover $r_\theta$ (i.e. $\Delta r_{\theta}(x,y_1,y_2)=\Delta r_1^*(x,y_1,y_2)+\Delta r_2^*(y_1,y_2)$).\\
To prove the existence of such $r_1^*$ and $r_2^*$, we first replace $\Delta r_1^*(x, y_1, y_2)$ with $\Delta r_\theta(x, y_1, y_2) - \Delta r_2^*(y_1, y_2)$. In the following proof, we have to use the following lemma:
\begin{lemma}
    $\mathbb{E}_{x\sim P(X|Y_1=y_1, Y_2=y_2)}[\sigma(\Delta r_\theta(x, y_1, y_2) - \Delta r_2^*(y_1, y_2))]$ \textit{monotonically decrease with the increase of }$\Delta r_2^*(y_1, y_2)$
\end{lemma}
\begin{proof}
    We directly take the derivative of $\mathbb{E}_{x\sim P(X|Y_1=y_1, Y_2=y_2)}[\sigma(\Delta r_\theta(x, y_1, y_2) - \Delta r_2^*(y_1, y_2))]$. Since $\Delta r_2^*(y_1, y_2))$ retains the same across different $x$, we have:
    \begin{align}
    &\frac{\partial [\mathbb{E}_{x\sim P(X|Y_1=y_1, Y_2=y_2)}[\sigma(\Delta r_\theta(x, y_1, y_2) - \Delta r_2^*(y_1, y_2))]]}{\partial \Delta r_2^*(y_1, y_2)} \\
    &= \frac{\mathbb{E}_{x\sim P(X|Y_1=y_1, Y_2=y_2)}[\partial \sigma(\Delta r_\theta(x, y_1, y_2) - \Delta r_2^*(y_1, y_2))]}{\partial \Delta r_2^*(y_1, y_2)} \\
        &= \mathbb{E}_{x\sim P(X|Y_1=y_1, Y_2=y_2)}[\frac{\partial \frac{1}{1+\exp(\Delta r_2^*(y_1, y_2)-\Delta r_\theta(x, y_1, y_2))}}{\partial (\Delta r_2^*(y_1, y_2)-\Delta r_\theta(x, y_1, y_2))}\cdot \frac{\partial (\Delta r_2^*(y_1, y_2)-\Delta r_\theta(x, y_1, y_2))}{\Delta r_2^*(y_1, y_2)}] \\
        &= \mathbb{E}_{x\sim P(X|Y_1=y_1, Y_2=y_2)}[-\frac{\exp(\Delta r_2^*(y_1, y_2)-\Delta r_\theta(x, y_1, y_2))}{(1 + \exp(\Delta r_2^*(y_1, y_2)-\Delta r_\theta(x, y_1, y_2)))^2}]
    \end{align}
    This shows that for any $y_1, y_2$, $\mathbb{E}_{x\sim P(X|Y_1=y_1, Y_2=y_2)}[\sigma(\Delta r_\theta(x, y_1, y_2) - \Delta r_2^*(y_1, y_2))]$ monotonically decrease with the increase of $\Delta r_2^*(y_1, y_2)$.
\end{proof}
To continue the proof of Theorem 2, we identify the lower and the upper bounds of $\mathbb{E}_{x\sim P(X|Y_1=y_1, Y_2=y_2)}[\sigma(\Delta r_\theta(x, y_1, y_2) - \Delta r_2^*(y_1, y_2))]$, under different choices of $\Delta r_2^*(y_1, y_2)$. Due to the assumption of bounded $r_\theta$, we consider $r_\theta$ that satisfies $\forall x, y_1, y_2, ~~ r_\theta(x, y_1, y_2) \in [r_{min}, r_{max}]$. The following holds:
\begin{align}
    &\mathbb{E}_{x\sim P(X|Y_1=y_1, Y_2=y_2)}[\sigma(\Delta r_\theta(x, y_1, y_2) - (r_{min}-r_{max}))] \\
    &=\mathbb{E}_{x\sim P(X|Y_1=y_1, Y_2=y_2)}[\sigma\Big( (r_\theta(x, y_1) - r_\theta(x, y_2)) - (r_{min}-r_{max})\Big)] \\
    &=\mathbb{E}_{x\sim P(X|Y_1=y_1, Y_2=y_2)}[\sigma\Big(\underbrace{(r_\theta(x, y_1)-r_{min})}_{\text{always $\ge$ 0}} + \underbrace{(r_{max}-r_\theta(x, y_2))}_{\text{always $\ge$ 0}}\Big)] \\
    &\ge \mathbb{E}_{x\sim P(X|Y_1=y_1, Y_2=y_2)}[\frac{1}{2}]=\frac{1}{2}
\end{align}
On the other hand, the following inequality also characterizes the lower bound of $\mathbb{E}_{x\sim P(X|Y_1=y_1, Y_2=y_2)}[\sigma(\Delta r_\theta(x, y_1, y_2) - \Delta r_2^*(y_1, y_2))]$:
\begin{align}
    &\mathbb{E}_{x\sim P(X|Y_1=y_1, Y_2=y_2)}[\sigma(\Delta r_\theta(x, y_1, y_2) - (r_{max}-r_{min}))] \\
    &=\mathbb{E}_{x\sim P(X|Y_1=y_1, Y_2=y_2)}[\sigma\Big( (r_\theta(x, y_1) - r_\theta(x, y_2)) - (r_{max}-r_{min})\Big)] \\
    &=\mathbb{E}_{x\sim P(X|Y_1=y_1, Y_2=y_2)}[\sigma\Big(\underbrace{(r_\theta(x, y_1)-r_{max})}_{\text{always $\le$ 0}} + \underbrace{(r_{min}-r_\theta(x, y_2))}_{\text{always $\le$ 0}}\Big)] \\
    &\le \mathbb{E}_{x\sim P(X|Y_1=y_1, Y_2=y_2)}[\frac{1}{2}]=\frac{1}{2}
\end{align}
Combine with the fact that $\mathbb{E}_{x\sim P(X|Y_1=y_1, Y_2=y_2)}[\sigma(\Delta r_\theta(x, y_1, y_2) - \Delta r_2^*(y_1, y_2))]$ monotonically decrease with the increase of $\Delta r_2^*(y_1, y_2)$, there always exists $\Delta r_2^*(y_1, y_2)$ such that $\mathbb{E}_{x\sim P(X|Y_1=y_1, Y_2=y_2)}[\sigma(\Delta r_\theta(x, y_1, y_2) - \Delta r_2^*(y_1, y_2))]=\frac{1}{2}$. To satisfy the feasibility condition that $\Delta r_{\theta}(x,y_1,y_2)=\Delta r_1(x,y_1,y_2)+\Delta r_2(y_1,y_2)$, we directly set $\Delta r_1^*(x,y_1,y_2)=\Delta r_{\theta}(x,y_1,y_2)-\Delta r_2^*(y_1,y_2)$, with the $\Delta r_2^*(y_1,y_2)$ obtained before. Such $\Delta r_2^*(y_1,y_2)$ still satisfies the simple structure mentioned in Theorem 1, so the constraint $\text{MI}(\tilde{Z}~\|~\tilde{W}) = \text{MI}(\tilde{Z}~\|~W)$ is still satisfied. The remaining task is to proof that the $\Delta r_1^*(x,y_1,y_2)$ obtained before satisfy the constraint $\text{MI}(Z~\|~\tilde{W})=0$. 

With the same proving technique in the proof of Theorem 1, we first decompose $\text{MI}(Z~\|~\tilde{W})$ into three entropy terms:
\begin{equation}
    \text{MI}(Z~\|~\tilde{W})=\text{H}({Z}) + \text{H}(\tilde{W}) - \text{H}({Z} , \tilde{W})
\end{equation}
We further expand the right side of the equation in the following equations:
\begin{align}
    \text{H}(Z) &= -Pr[Z=1]\log(Pr[Z=1])-Pr[Z=0]\log(Pr[Z=0]) \nonumber \\
    \text{H}(\tilde{W}) &= -Pr[\tilde{W}=1]\log(Pr[\tilde{W}=1])-Pr[\tilde{W}=0]\log(Pr[\tilde{W}=0]) \nonumber \\
    \text{H}({Z} , \tilde{W})&= -Pr[{Z}=1, \tilde{W}=1]\log(Pr[{Z}=1, \tilde{W}=1]) - Pr[{Z}=1, \tilde{W}=0]\log(Pr[{Z}=1, \tilde{W}=0]) \nonumber\\
    &- Pr[{Z}=0, \tilde{W}=1]\log(Pr[{Z}=0, \tilde{W}=1]) - Pr[{Z}=0, \tilde{W}=0]\log(Pr[{Z}=0, \tilde{W}=0])
\end{align}
With simple algebraic simplification, we have:
\begin{align}
    &\text{H}({Z}) + \text{H}(\tilde{W}) - \text{H}({Z} , \tilde{W}) \nonumber \\
    &= Pr[{Z}=1, \tilde{W}=1]\log(Pr[{Z}=1, \tilde{W}=1])+Pr[{Z}=1, \tilde{W}=0]\log(Pr[{Z}=1, \tilde{W}=0]) \nonumber \\
    &+Pr[{Z}=0, \tilde{W}=1]\log(Pr[{Z}=0, \tilde{W}=1])+Pr[{Z}=0, \tilde{W}=0]\log(Pr[{Z}=0, \tilde{W}=0])\nonumber\\
    &-Pr[Z=1]\log (Pr[Z=1])-Pr[Z=0]\log (Pr[Z=0])-Pr[\tilde{W}=1]\log(\tilde{W}=1)\nonumber\\
    &-Pr[\tilde{W}=0]\log(\tilde{W}=0) 
\end{align}
To prove that $\text{MI}(Z~\|~\tilde{W})=\text{H}({Z}) + \text{H}(\tilde{W}) - \text{H}({Z} , \tilde{W})=0$, we first expand the inner probability with the same technique use in proving Theorem 1, and then perform the following transformation for $\mathbb{E}_{\text{Ber}}[\mathbb{I}[{Z}]]$ and $\mathbb{E}_{\text{Ber}}[\mathbb{I}[\tilde{W}]]$:
\begin{align}
    \mathbb{E}_{\text{Ber}}[\mathbb{I}[{Z}]]=\mathbb{E}_{\text{Ber}}[\mathbb{I}[{Z}]]\cdot\Big(\mathbb{E}_{\text{Ber}}[\mathbb{I}[\tilde{W}]] + (1 - \mathbb{E}_{\text{Ber}}[\mathbb{I}[\tilde{W}]])\Big) \\
    \mathbb{E}_{\text{Ber}}[\mathbb{I}[\tilde{W}]]=\mathbb{E}_{\text{Ber}}[\mathbb{I}[\tilde{W}]]\cdot\Big(\mathbb{E}_{\text{Ber}}[\mathbb{I}[{Z}]] + (1 - \mathbb{E}_{\text{Ber}}[\mathbb{I}[{Z}]])\Big)
\end{align}
Note that this simple transformation bridges all terms. Take the calculation of $Pr[{Z}=1, \tilde{W}=1]\log(Pr[{Z}=1, \tilde{W}=1])-Pr[Z=1]\log (Pr[Z=1])-Pr[\tilde{W}=1]\log(\tilde{W}=1) $ for an example, we expand the inner probability with the same technique use in proving Theorem 1:
\begin{align}
    &Pr[{Z}=1, \tilde{W}=1]\log(Pr[{Z}=1, \tilde{W}=1])-Pr[Z=1]\log (Pr[Z=1])-Pr[\tilde{W}=1]\log(\tilde{W}=1) \\
    &=\mathbb{E}_{P(Y_1, Y_2)}\mathbb{E}_{P(X|Y_1, Y_2)} [\mathbb{E}_{\text{Ber}\times\text{Ber}}[\mathbb{I}[Z\cdot \tilde{W}]]] \log (\mathbb{E}_{P(Y_1, Y_2)}\mathbb{E}_{P(X|Y_1, Y_2)} [\mathbb{E}_{\text{Ber}\times\text{Ber}}[\mathbb{I}[Z\cdot \tilde{W}]]]) \nonumber \\
    & - \mathbb{E}_{P(Y_1, Y_2)}\mathbb{E}_{P(X|Y_1, Y_2)}[\mathbb{E}_{\text{Ber}}[\mathbb{I}[{Z}]]]\log(\mathbb{E}_{P(Y_1, Y_2)}\mathbb{E}_{P(X|Y_1, Y_2)}[\mathbb{E}_{\text{Ber}}[\mathbb{I}[{Z}]]]) \nonumber \\
    & - \mathbb{E}_{P(Y_1, Y_2)}\mathbb{E}_{P(X|Y_1, Y_2)}[\mathbb{E}_{\text{Ber}}[\mathbb{I}[\tilde{W}]]]\log(\mathbb{E}_{P(Y_1, Y_2)}\mathbb{E}_{P(X|Y_1, Y_2)}[\mathbb{E}_{\text{Ber}}[\mathbb{I}[\tilde{W}]]])
\end{align}
After using the previously mentioned transformation, we have the right-hand-side above equals to:
\begin{align}
    &\mathbb{E}_{P(Y_1, Y_2)}\mathbb{E}_{P(X|Y_1, Y_2)} [\mathbb{E}_{\text{Ber}\times\text{Ber}}[\mathbb{I}[Z\cdot \tilde{W}]]] \log (\mathbb{E}_{P(Y_1, Y_2)}\mathbb{E}_{P(X|Y_1, Y_2)} [\mathbb{E}_{\text{Ber}\times\text{Ber}}[\mathbb{I}[Z\cdot \tilde{W}]]]) \nonumber \\
    &-\mathbb{E}_{P(Y_1, Y_2)}\mathbb{E}_{P(X|Y_1, Y_2)} \Big[\mathbb{E}_{\text{Ber}}[\mathbb{I}[{Z}]]\cdot\mathbb{E}_{\text{Ber}}[\mathbb{I}[\tilde{W}]]\Big]\ast \nonumber\\
    &~~~~~~~~~~~~~~~~~~~~~~~~~~~~~~~\log\Big(\mathbb{E}_{P(Y_1, Y_2)}\mathbb{E}_{P(X|Y_1, Y_2)}[\mathbb{E}_{\text{Ber}}[\mathbb{I}[{Z}]]]\cdot \mathbb{E}_{P(Y_1, Y_2)}\mathbb{E}_{P(X|Y_1, Y_2)}[\mathbb{E}_{\text{Ber}}[\mathbb{I}[\tilde{W}]]]\Big) \nonumber\\
    &-\underbrace{\mathbb{E}_{P(Y_1, Y_2)}\mathbb{E}_{P(X|Y_1, Y_2)} \Big[\mathbb{E}_{\text{Ber}}[\mathbb{I}[{Z}]]\cdot(1-\mathbb{E}_{\text{Ber}}[\mathbb{I}[\tilde{W}]])\Big]\log(\mathbb{E}_{P(Y_1, Y_2)}\mathbb{E}_{P(X|Y_1, Y_2)}[\mathbb{E}_{\text{Ber}}[\mathbb{I}[{Z}]]])}_{\text{used for symmetric construction for canceling other term}} \nonumber \\
    &- \underbrace{\mathbb{E}_{P(Y_1, Y_2)}\mathbb{E}_{P(X|Y_1, Y_2)} \Big[(1-\mathbb{E}_{\text{Ber}}[\mathbb{I}[{Z}]])\cdot\mathbb{E}_{\text{Ber}}[\mathbb{I}[{W}]]\Big]\log(\mathbb{E}_{P(Y_1, Y_2)}\mathbb{E}_{P(X|Y_1, Y_2)}[\mathbb{E}_{\text{Ber}}[\mathbb{I}[\tilde{W}]]])}_{\text{used for symmetric construction for canceling other term}} 
\end{align}
With simple verification, we can find that the last two terms can be used to cancel other terms that come from $\text{H}({Z} , \tilde{W})$. This means we only have to prove that the first two terms in the above equation cancel out. In other words, all we have to prove is:
\begin{align}
    &\mathbb{E}_{P(Y_1, Y_2)}\mathbb{E}_{P(X|Y_1, Y_2)} [\mathbb{E}_{\text{Ber}\times\text{Ber}}[\mathbb{I}[Z\cdot \tilde{W}]]] \log (\mathbb{E}_{P(Y_1, Y_2)}\mathbb{E}_{P(X|Y_1, Y_2)} [\mathbb{E}_{\text{Ber}\times\text{Ber}}[\mathbb{I}[Z\cdot \tilde{W}]]]) \nonumber \\
    &-\mathbb{E}_{P(Y_1, Y_2)}\mathbb{E}_{P(X|Y_1, Y_2)} \Big[\mathbb{E}_{\text{Ber}}[\mathbb{I}[{Z}]]\cdot\mathbb{E}_{\text{Ber}}[\mathbb{I}[\tilde{W}]]\Big]\ast \nonumber\\
    &~~~~~~~~~~~~~~~~~~~~~~~~~~~~~~~~~~\log\Big(\mathbb{E}_{P(Y_1, Y_2)}\mathbb{E}_{P(X|Y_1, Y_2)}[\mathbb{E}_{\text{Ber}}[\mathbb{I}[{Z}]]]\cdot \mathbb{E}_{P(Y_1, Y_2)}\mathbb{E}_{P(X|Y_1, Y_2)}[\mathbb{E}_{\text{Ber}}[\mathbb{I}[\tilde{W}]]]\Big) \nonumber\\
    &=\mathbb{E}_{P(Y_1, Y_2)}\mathbb{E}_{P(X|Y_1, Y_2)} \Big[\mathbb{E}_{\text{Ber}}[\mathbb{I}[{Z}]]\cdot\mathbb{E}_{\text{Ber}}[\mathbb{I}[\tilde{W}]]\Big]\ast \nonumber\\
    &~~~~~~~~~~~~~~~~~~~~~~~~~~~~~~~\log \Bigg( \frac{\mathbb{E}_{P(Y_1, Y_2)}\mathbb{E}_{P(X|Y_1, Y_2)} \Big[\mathbb{E}_{\text{Ber}}[\mathbb{I}[{Z}]]\cdot\mathbb{E}_{\text{Ber}}[\mathbb{I}[\tilde{W}]]\Big]}{\mathbb{E}_{P(Y_1, Y_2)}\mathbb{E}_{P(X|Y_1, Y_2)}[\mathbb{E}_{\text{Ber}}[\mathbb{I}[{Z}]]]\cdot \mathbb{E}_{P(Y_1, Y_2)}\mathbb{E}_{P(X|Y_1, Y_2)}[\mathbb{E}_{\text{Ber}}[\mathbb{I}[\tilde{W}]]]} \Bigg) \nonumber \\
    &=0
\end{align}
Note that we change all $\mathbb{E}_{P(Y_1, Y_2)}\mathbb{E}_{P(X|Y_1, Y_2)} [\mathbb{E}_{\text{Ber}\times\text{Ber}}[\mathbb{I}[Z\cdot \tilde{W}]]]$ into $\mathbb{E}_{P(Y_1, Y_2)}\mathbb{E}_{P(X|Y_1, Y_2)} \Big[\mathbb{E}_{\text{Ber}}[\mathbb{I}[{Z}]]\cdot\mathbb{E}_{\text{Ber}}[\mathbb{I}[\tilde{W}]]\Big]$ due to independency similar as before. We can finally use the property that for any $y_1, y_2$, $\mathbb{E}_{x\sim P(X|Y_1=y_1, Y_2=y_2)}[\sigma(\Delta r_\theta(x, y_1, y_2) - \Delta r_2^*(y_1, y_2))]=\frac{1}{2}$, and remember that $\mathbb{E}_{\text{Ber}}[\mathbb{I}[\tilde{W}]]$ is independent of specific $x$, then we can perform the following simplification:
\begin{align}
    &\mathbb{E}_{P(Y_1, Y_2)}\mathbb{E}_{P(X|Y_1, Y_2)} \Big[\mathbb{E}_{\text{Ber}}[\mathbb{I}[{Z}]]\cdot\mathbb{E}_{\text{Ber}}[\mathbb{I}[\tilde{W}]]\Big] \ast \nonumber \\
    &~~~~~~~~~~~~~~~~~~~~~~~~~~~~~~~~~\log \Bigg( \frac{\mathbb{E}_{P(Y_1, Y_2)}\mathbb{E}_{P(X|Y_1, Y_2)} \Big[\mathbb{E}_{\text{Ber}}[\mathbb{I}[{Z}]]\cdot\mathbb{E}_{\text{Ber}}[\mathbb{I}[\tilde{W}]]\Big]}{\mathbb{E}_{P(Y_1, Y_2)}\mathbb{E}_{P(X|Y_1, Y_2)}[\mathbb{E}_{\text{Ber}}[\mathbb{I}[{Z}]]]\cdot \mathbb{E}_{P(Y_1, Y_2)}\mathbb{E}_{P(X|Y_1, Y_2)}[\mathbb{E}_{\text{Ber}}[\mathbb{I}[\tilde{W}]]]} \Bigg) \nonumber \\
    &=\mathbb{E}_{P(Y_1, Y_2)}\mathbb{E}_{P(X|Y_1, Y_2)} \Big[\mathbb{E}_{\text{Ber}}[\mathbb{I}[{Z}]]\cdot\mathbb{E}_{\text{Ber}}[\mathbb{I}[\tilde{W}]]\Big]\ast \nonumber \\
    &~~~~~~~~~~~~~~~~~~~~~~~~~~~~~~~~~~~~~\log \Bigg( \frac{\mathbb{E}_{P(Y_1, Y_2)}\Big[\Big(\mathbb{E}_{P(X|Y_1, Y_2)} \mathbb{E}_{\text{Ber}}[\mathbb{I}[{Z}]]\Big)\cdot\mathbb{E}_{\text{Ber}}[\mathbb{I}[\tilde{W}]]\Big]}{\mathbb{E}_{P(Y_1, Y_2)}\Big(\mathbb{E}_{P(X|Y_1, Y_2)}[\mathbb{E}_{\text{Ber}}[\mathbb{I}[{Z}]]]\Big)\cdot \mathbb{E}_{P(Y_1, Y_2)}[\mathbb{E}_{\text{Ber}}[\mathbb{I}[\tilde{W}]]]} \Bigg) \nonumber \\
    &=\mathbb{E}_{P(Y_1, Y_2)}\mathbb{E}_{P(X|Y_1, Y_2)} \Big[\mathbb{E}_{\text{Ber}}[\mathbb{I}[{Z}]]\cdot\mathbb{E}_{\text{Ber}}[\mathbb{I}[\tilde{W}]]\Big]\log \Bigg( \frac{\mathbb{E}_{P(Y_1, Y_2)}\Big[\Big(\frac{1}{2}\Big)\cdot\mathbb{E}_{\text{Ber}}[\mathbb{I}[\tilde{W}]]\Big]}{\mathbb{E}_{P(Y_1, Y_2)}\Big(\frac{1}{2}\Big)\cdot \mathbb{E}_{P(Y_1, Y_2)}[\mathbb{E}_{\text{Ber}}[\mathbb{I}[\tilde{W}]]]} \Bigg) \nonumber \\
    &=\mathbb{E}_{P(Y_1, Y_2)}\mathbb{E}_{P(X|Y_1, Y_2)} \Big[\mathbb{E}_{\text{Ber}}[\mathbb{I}[{Z}]]\cdot\mathbb{E}_{\text{Ber}}[\mathbb{I}[\tilde{W}]]\Big]\log (\frac{\frac{1}{2}}{\frac{1}{2}}) =0
\end{align}
After verifying the feasibility of the solution, we verify that such $r_1^*$ and $r_2^*$ is optimal for maximizing $\text{H}(Z)$. By definition, we have:
\begin{align}
    \text{H}(Z)=-Pr[Z=1]\log (Pr[Z=1]) - Pr[Z=0]\log (Pr[Z=0])
\end{align}
To calculate $Pr[Z=1]$, we can use the same expansion technique before. We have:
\begin{align}
    Pr[Z=1]&=\mathbb{E}_{P(Y_1, Y_2)}\mathbb{E}_{P(X|Y_1, Y_2)} [\mathbb{E}_{\text{Ber}}[\mathbb{I}[{Z}]]] \nonumber \\
    &= \sum_{y_1, y_2}P(Y_1=y_1, Y_2=y_2)[\sum_{x}P(X=x|Y_1=y_1, Y_2=y_2)[\mathbb{E}_{\text{Ber}}[\mathbb{I}[Z]]] \nonumber \\
    &= \sum_{y_1, y_2}P(Y_1=y_1, Y_2=y_2)[\sum_{x}P(X=x|Y_1=y_1, Y_2=y_2)[\sigma(\Delta r_1^*(x, y_1, y_2))]] \nonumber \\
    &=  \sum_{y_1, y_2}P(Y_1=y_1, Y_2=y_2)[\mathbb{E}_{x\sim P(X|Y_1=y_1, Y_2=y_2)}[\sigma(\Delta r_1^*(x, y_1, y_2))]=\frac{1}{2}
\end{align}
It's easy to verify that such $r_1^*$ and $r_2^*$ reach the maximum value of $\text{H}(Z)$, which is the optimal solution we want.
\end{proof}
\section{More about experiments}
\subsection{Full results in Table \ref{tab: length-biased and adversarial} and Table \ref{tab:combined-results}} 
\label{app: full results}
Due to space limits, we only reported mean scores from 3 random seeds in the main context. Full results (mean and std) in Table \ref{tab: length-biased and adversarial} and Table \ref{tab:combined-results} are listed below, showing consistent performance among different seeds.

\begin{table}[H]
    \small
    \centering
    \begin{tabular}{cccccc}
            \hline
            Model & Chat & Chat Hard & Safety & Reason & Avg \\
            \hline
            vanilla & 78.0\textcolor{gray}{$\pm$2.1} & 29.8\textcolor{gray}{$\pm$0.7} & 36.4$\pm$\textcolor{gray}{0.9} & 58.2$\pm$\textcolor{gray}{1.1} & 50.6$\pm$\textcolor{gray}{1.0} \\
            \\
            ours    & 86.8$\pm$\textcolor{gray}{1.8} & 31.1$\pm$\textcolor{gray}{0.8} & 45.1$\pm$\textcolor{gray}{0.8} & 60.3$\pm$\textcolor{gray}{1.2} & 55.8$\pm$\textcolor{gray}{1.1} \\
            \hline
        \end{tabular}
    \caption{Full results of experiments in Table \ref{tab: length-biased and adversarial} (a). We conduct the same experiments with 3 different random seeds and report the mean and std values of the results.}
\end{table}

\begin{table}[H]
    \small
    \centering
    \begin{tabular}{cccccc}
            \hline
            Model & Chat & Chat Hard & Safety & Reason & Avg \\
            \hline
            vanilla  & 80.7\textcolor{gray}{$\pm$2.1} & 28.5\textcolor{gray}{$\pm$0.8} & 40.2\textcolor{gray}{$\pm$1.0} & 36.5\textcolor{gray}{$\pm$1.5} & 46.4\textcolor{gray}{$\pm$1.3} \\
            original & 84.9\textcolor{gray}{$\pm$1.7} & 31.3\textcolor{gray}{$\pm$0.6} & 42.8\textcolor{gray}{$\pm$1.1} & 49.0\textcolor{gray}{$\pm$1.7} & 52.0\textcolor{gray}{$\pm$0.9} \\
            ours     & 84.3\textcolor{gray}{$\pm$1.8} & 30.9\textcolor{gray}{$\pm$0.5} & 43.2\textcolor{gray}{$\pm$0.9} & 46.7\textcolor{gray}{$\pm$1.0} & 51.6\textcolor{gray}{$\pm$0.8} \\
            \hline
        \end{tabular}
    \caption{Full results of experiments in Table \ref{tab: length-biased and adversarial} (b). We conduct the same experiments with 3 different random seeds and report the mean and std values of the results.}
\end{table}

\begin{table}[H]
\small
\centering
\begin{tabular}{cccccc}
\hline
Method & Chat & Chat Hard & Safety & Reasoning & Average \\
\hline
vanilla-8B & 0.93\textcolor{gray}{$\pm$0.01} & 0.50\textcolor{gray}{$\pm$0.02} & 0.67\textcolor{gray}{$\pm$0.02} & 0.78\textcolor{gray}{$\pm$0.03} & 0.72\textcolor{gray}{$\pm$0.02} \\
RRM-8B & 0.95\textcolor{gray}{$\pm$0.01} & 0.56\textcolor{gray}{$\pm$0.01} & 0.75\textcolor{gray}{$\pm$0.02} & 0.82\textcolor{gray}{$\pm$0.02} & 0.77\textcolor{gray}{$\pm$0.01} \\
Ours-8B & \textbf{0.96}\textcolor{gray}{$\pm$0.02} & \textbf{0.59}\textcolor{gray}{$\pm$0.01} & \textbf{0.81}\textcolor{gray}{$\pm$0.01} & \textbf{0.89}\textcolor{gray}{$\pm$0.02} & \textbf{0.82}\textcolor{gray}{$\pm$0.01} \\
vanilla-7B & 0.90\textcolor{gray}{$\pm$0.02} & 0.49\textcolor{gray}{$\pm$0.03} & 0.60\textcolor{gray}{$\pm$0.01} & 0.69\textcolor{gray}{$\pm$0.05} & 0.66\textcolor{gray}{$\pm$0.03} \\
RRM-7B & 0.94\textcolor{gray}{$\pm$0.01} & 0.52\textcolor{gray}{$\pm$0.01} & 0.65\textcolor{gray}{$\pm$0.01} & 0.70\textcolor{gray}{$\pm$0.03} & 0.70\textcolor{gray}{$\pm$0.01} \\
Ours-7B & \textbf{0.94}\textcolor{gray}{$\pm$0.01} & \textbf{0.56}\textcolor{gray}{$\pm$0.02} & \textbf{0.69}\textcolor{gray}{$\pm$0.02} & \textbf{0.81}\textcolor{gray}{$\pm$0.04} & \textbf{0.75}\textcolor{gray}{$\pm$0.02} \\
\hline
\end{tabular}
\caption{Full results of experiments in Table \ref{tab:combined-results} (a). We conduct the same experiments with 3 different random seeds and report the mean and std values of the results.}
\label{tab:reward_bench}
\end{table}

\begin{table}[H]
\small
\centering
\begin{tabular}{cccc}
\hline
Method & vanilla & RRM & Ours \\
\hline
DPO & 30.5\textcolor{gray}{$\pm$1.6} / 38.8\textcolor{gray}{$\pm$1.3} & 39.5\textcolor{gray}{$\pm$0.6} / 40.9\textcolor{gray}{$\pm$0.8} & \textbf{40.6}\textcolor{gray}{$\pm$1.2} / \textbf{42.3}\textcolor{gray}{$\pm$0.9} \\
BoN(N=4) & 33.2\textcolor{gray}{$\pm$2.1} / 42.5\textcolor{gray}{$\pm$1.7} & 35.3\textcolor{gray}{$\pm$1.4} / 38.4\textcolor{gray}{$\pm$1.5} & \textbf{38.5}\textcolor{gray}{$\pm$1.0} / \textbf{45.0}\textcolor{gray}{$\pm$1.1} \\
BoN(N=32) & 35.8\textcolor{gray}{$\pm$1.8} / 45.1\textcolor{gray}{$\pm$2.0} & 41.1\textcolor{gray}{$\pm$1.9} / 44.9\textcolor{gray}{$\pm$1.2} & \textbf{44.7}\textcolor{gray}{$\pm$1.7} / \textbf{48.3}\textcolor{gray}{$\pm$0.9} \\
\hline
\end{tabular}
\caption{Full AlpacaEval-2 (LCWR / WR) results. We conduct the same experiments with 3 different random seeds and report the mean and std values of the results.}
\label{tab:alpacaeval2}
\end{table}

\begin{table}[H]
\small
\centering
\begin{tabular}{cccc}
\hline
Method & vanilla & RRM & Ours \\
\hline
DPO & 7.83$\pm$0.12 / 6.51$\pm$0.30 & 8.35$\pm$0.15 / 7.46$\pm$0.24 & \textbf{8.44}\textcolor{gray}{$\pm$0.10} / \textbf{8.03}\textcolor{gray}{$\pm$0.26} \\
\hline
\end{tabular}
\caption{Full MT-Bench (T1 / T2) results. We conduct the same experiments with 3 different random seeds and report the mean and std values of the results.}
\label{tab:mt_bench}
\end{table}

\subsection{Reward-Bench results based on SHP dataset}
\label{app: Reward-Bench results based on  SHP dataset}

\begin{table}[H]
\small
\centering
\begin{tabular}{cccccc}
\hline
Method & Chat & Chat Hard & Safety & Reasoning & Average \\
\hline
vanilla-8B & 0.91\textcolor{gray}{$\pm$0.02} & 0.39\textcolor{gray}{$\pm$0.01} & 0.46\textcolor{gray}{$\pm$0.01} & 0.75\textcolor{gray}{$\pm$0.02} & 0.63\textcolor{gray}{$\pm$0.01} \\
RRM-8B & 0.93\textcolor{gray}{$\pm$0.02} & 0.45\textcolor{gray}{$\pm$0.01} & 0.54\textcolor{gray}{$\pm$0.03} & 0.77\textcolor{gray}{$\pm$0.01} & 0.67\textcolor{gray}{$\pm$0.02} \\
Ours-8B & \textbf{0.95}\textcolor{gray}{$\pm$0.01} & \textbf{0.46}\textcolor{gray}{$\pm$0.01} & \textbf{0.61}\textcolor{gray}{$\pm$0.01} & \textbf{0.88}\textcolor{gray}{$\pm$0.02} & \textbf{0.72}\textcolor{gray}{$\pm$0.01}\\
vanilla-7B & 0.84\textcolor{gray}{$\pm$0.02} & 0.33\textcolor{gray}{$\pm$0.03} & 0.45\textcolor{gray}{$\pm$0.05} & 0.59\textcolor{gray}{$\pm$0.03} & 0.55\textcolor{gray}{$\pm$0.04} \\
RRM-7B & 0.89\textcolor{gray}{$\pm$0.02} & 0.37\textcolor{gray}{$\pm$0.01} & 0.48\textcolor{gray}{$\pm$0.01} & 0.58\textcolor{gray}{$\pm$0.04} & 0.58\textcolor{gray}{$\pm$0.02} \\
Ours-7B & \textbf{0.90}\textcolor{gray}{$\pm$0.01} & \textbf{0.39}\textcolor{gray}{$\pm$0.01} & \textbf{0.52}\textcolor{gray}{$\pm$0.01} & \textbf{0.74}\textcolor{gray}{$\pm$0.03} & \textbf{0.64}\textcolor{gray}{$\pm$0.02} \\
\hline
\end{tabular}
\caption{Results on Reward-Bench: These experiments are based on the same two reward backbones, but with the SHP dataset. The improvement brought by our method is also significant on the SHP dataset. However, since the SHP dataset contains many fewer data samples and may inherently have more spurious preferences, the accuracy generally decreases.}
\label{tab:reward_bench_table3}
\end{table}

\subsection{Justification and analysis of binary clustering and EMA threshold used in data prioritization}
\label{app: Justification and analysis data prioritization}
An adaptive threshold is required since reward learning is dynamic—Figure \ref{fig:toy_examples} show that even small models (1B) experience significant $\Delta r_2$ scale changes during training.

During optimization, the reward model is updated in batches, with $\Delta r_2$ calculated from limited samples each step. Determining whether a sample's $\Delta r_2$ is "relatively small" in a batch becomes a 1D binary-clustering problem($\Delta r_2$ as feature). Due to high variance in $\Delta r_2$ values among different samples (as shown in Figure \ref{fig:toy_examples}), relying solely on current batch data is unstable and cannot truly reflect the properties of $\Delta r_2$. Our dynamic threshold, computed via exponential moving average, incorporates recent training steps, dynamically capturing $\Delta r_2$ properties while reducing threshold variance.

As for the robustness, we conduct the same experiments on Reward-Bench in Section 4.2 with different EMA weights ($\alpha$ in Alg. \ref{alg: select sample with context-free reward}) and clustering methods (Otsu, K-means). Results show robustness to thresholding methods. Limited by time and budget, we only use LLaMA-3-8B-Instruct and SHP dataset. We believe the robustness partially stems from the fact that samples with larger $\Delta r_2$ in the current batch will not be directly discarded, preserving the chance to be used in the future step.

\begin{table}[h]
\centering
\small
\begin{tabular}{c|c|c|c|c|c|c}
\hline
Method & $\alpha$ & Chat & Chat Hard & Safety & Reasoning & Average \\
\hline
Otsu     & 0.9 & 0.95 & 0.48 & 0.59 & 0.87 & 0.72 \\
Otsu     & 0.8 & 0.92 & 0.49 & 0.57 & 0.84 & 0.70 \\
K-means  & 0.9 & 0.95 & 0.46 & 0.61 & 0.88 & 0.72 \\
K-means  & 0.8 & 0.93 & 0.48 & 0.60 & 0.87 & 0.71 \\
\hline
\end{tabular}
\caption{Performance under different methods and $\alpha$ values.}
\label{tab:method_alpha_performance}
\end{table}

\subsection{experimental details}
\label{app: experimental details}
Our implementation is based on the OpenRLHF \cite{hu2024openrlhf} framework, which uses Apache-2.0 license. Experiments are run on Nvidia A100(40G) GPUs. For reward training, we use 8*A100(40G) GPUs so that the training can be finished in 12 hours. For DPO training and response generation, we also use 8*A100(40G) GPUs, and the time consumption is similar to the reward training. All the reward models and the LLMs are fully fine-tuned. We use Deepspeed \cite{aminabadi2022deepspeed} as the framework of parallelization. For the implementation of RRM, we reference its official code in RLHFlow\footnote{https://github.com/RLHFlow}. We use an AdamW optimizer with a learning rating of 5e-7 and 9e-6 for the DPO policy and the reward model, respectively. For other general hyperparameters, we follow the default parameters in OpenRLHF.

Our exclusive hyperparameters include clustering methods and EMA weights for $\Delta r_2$ thresholding. In practice, we tried Otsu's method and K-means for clustering method and 0.8, 0.9 for EMA weight. As is shown in Appendix \ref{app: Justification and analysis data prioritization}, our method performs consistently, so we select the best-performing hyperparameters. 

We continue to explain the data construction process for the toy cases. For the length-biased dataset, we use all preference pairs $(x, y_w, y_l)$ where the chosen response is longer, i.e., $|y_w| > |y_l|$, from the original dataset; based on the number of chosen-longer samples, we select other samples with $\frac{1}{4}$ number of them, whose chosen response is shorter. The combined subsets create a clearly length-biased dataset. For the adversarial prompt dataset, we use the following diagram to illustrate the data processing:
\begin{figure}[h] 
    \centering
    \begin{minipage}{0.48\textwidth} 
        \centering
        \includegraphics[width=\textwidth]{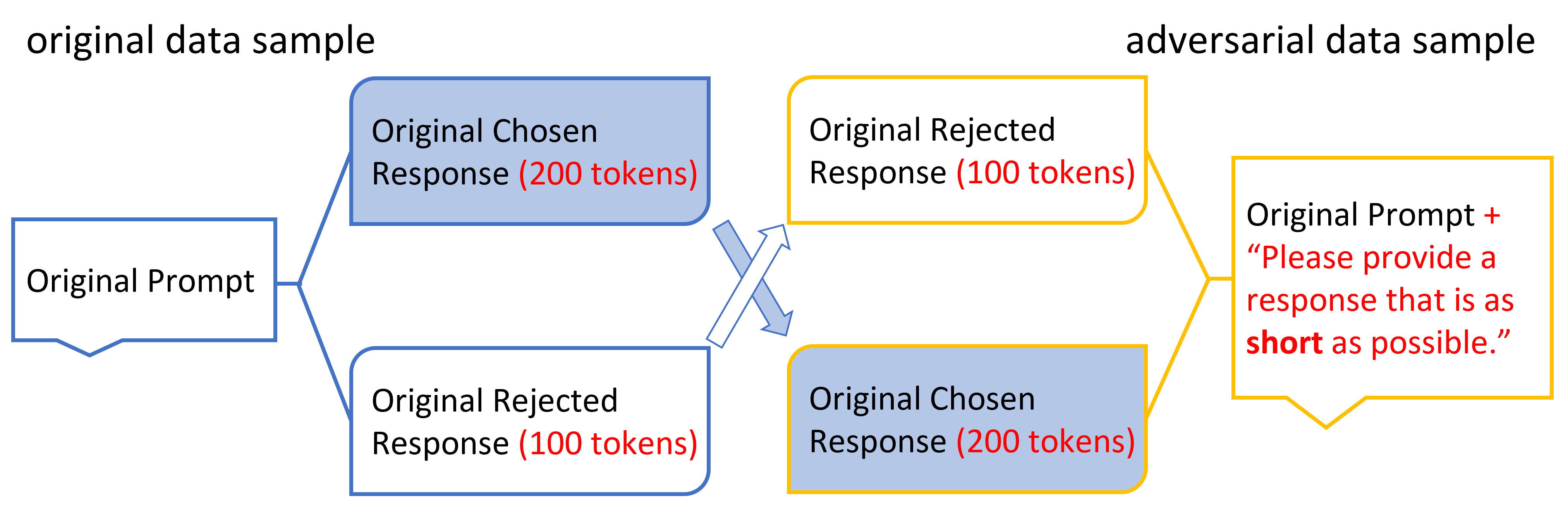} 
    \end{minipage}\hfill
    \begin{minipage}{0.48\textwidth} 
        \centering
        \includegraphics[width=\textwidth]{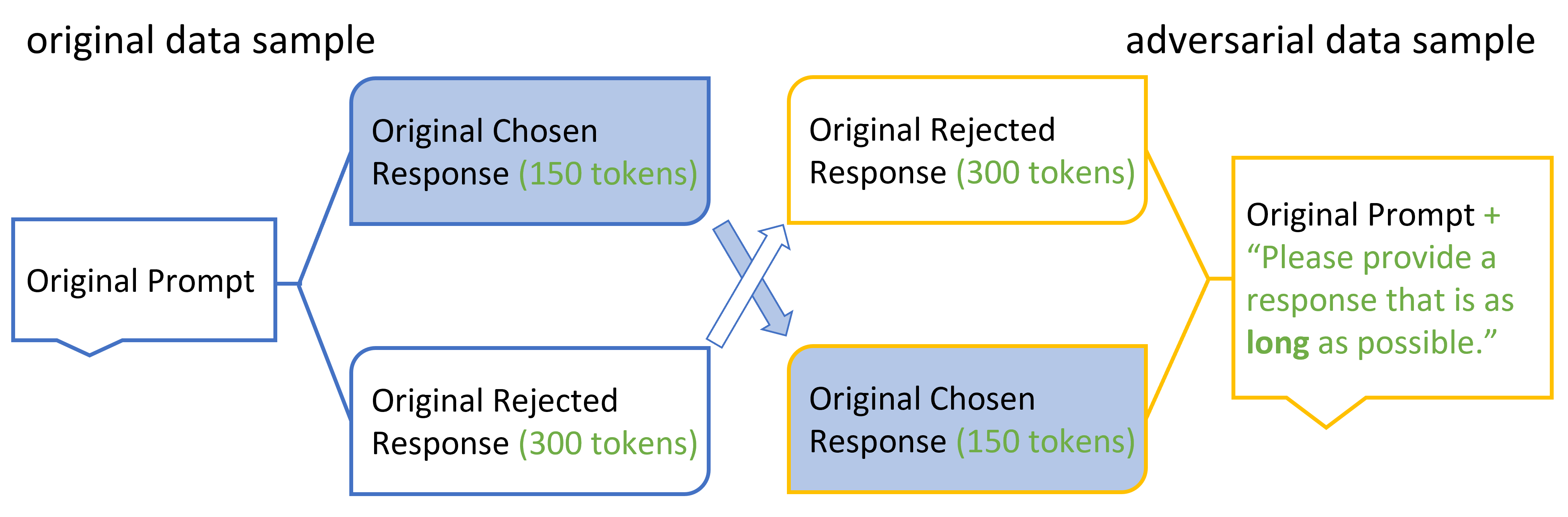} 
    \end{minipage}
\end{figure}

We note that RLHFLow uses Apache-2.0 license. SHP dataset didn't claim the license it uses. LLaMA models use Meta Llama 3 Community License Agreement. Mistral models use Apache-2.0 license.

\subsection{Explanations of the superiority compared with baseline methods}
\label{app: Explanations of the superiority compared with baseline methods}
The advantage of our method compared with vanilla training is obvious. Since vanilla training will use all data samples, including those samples with strong prompt-free preference. Thus, the resulting reward model easily overfits to some prompt-free preference existing in the dataset. Such prompt-free preference harms the reward model's generalization capability and leads to a low-quality alignment.

Compared with RRM, although both RRM and our method mitigate spurious preferences by altering the data distribution, RRM introduces hand-crafted but mismatched prompt-response pairs into the dataset and updates the reward model on them, which in turn affects the meaningful preferences. In contrast, our method leverages them solely to characterize the reward model during training and only trains on the original preference data, where each prompt-response pair is labeled by human preference. This contributes to the superior performance of our method compared to RRM. \\

\section{Boarder Impacts}
\label{Boarder Impacts}
This paper presents work whose goal is to advance the field of Machine Learning. There are many potential societal consequences of our work. For positive ones, our method could result in a more generalizable reward model and thus advance other RLHF methods. This can make the LLMs more aligned with human preferences. For negative ones, one can achieve jailbreaking in LLMs through malicious reward models.


\newpage
\section*{NeurIPS Paper Checklist}

The checklist is designed to encourage best practices for responsible machine learning research, addressing issues of reproducibility, transparency, research ethics, and societal impact. Do not remove the checklist: {\bf The papers not including the checklist will be desk rejected.} The checklist should follow the references and follow the (optional) supplemental material.  The checklist does NOT count towards the page
limit. 

Please read the checklist guidelines carefully for information on how to answer these questions. For each question in the checklist:
\begin{itemize}
    \item You should answer \answerYes{}, \answerNo{}, or \answerNA{}.
    \item \answerNA{} means either that the question is Not Applicable for that particular paper or the relevant information is Not Available.
    \item Please provide a short (1–2 sentence) justification right after your answer (even for NA). 
\end{itemize}

{\bf The checklist answers are an integral part of your paper submission.} They are visible to the reviewers, area chairs, senior area chairs, and ethics reviewers. You will be asked to also include it (after eventual revisions) with the final version of your paper, and its final version will be published with the paper.

The reviewers of your paper will be asked to use the checklist as one of the factors in their evaluation. While "\answerYes{}" is generally preferable to "\answerNo{}", it is perfectly acceptable to answer "\answerNo{}" provided a proper justification is given (e.g., "error bars are not reported because it would be too computationally expensive" or "we were unable to find the license for the dataset we used"). In general, answering "\answerNo{}" or "\answerNA{}" is not grounds for rejection. While the questions are phrased in a binary way, we acknowledge that the true answer is often more nuanced, so please just use your best judgment and write a justification to elaborate. All supporting evidence can appear either in the main paper or the supplemental material, provided in appendix. If you answer \answerYes{} to a question, in the justification please point to the section(s) where related material for the question can be found.

IMPORTANT, please:
\begin{itemize}
    \item {\bf Delete this instruction block, but keep the section heading ``NeurIPS Paper Checklist"},
    \item  {\bf Keep the checklist subsection headings, questions/answers and guidelines below.}
    \item {\bf Do not modify the questions and only use the provided macros for your answers}.
\end{itemize}


\begin{enumerate}

\item {\bf Claims}
    \item[] Question: Do the main claims made in the abstract and introduction accurately reflect the paper's contributions and scope?
    \item[] Answer: \answerYes{} 
    \item[] Justification: Our main claims can be reflected by Section \ref{section: method} and Section \ref{section: experiments}.

\item {\bf Limitations}
    \item[] Question: Does the paper discuss the limitations of the work performed by the authors?
    \item[] Answer: \answerYes{} 
    \item[] Justification: Limitations are discussed in Section \ref{section: conclusion}

\item {\bf Theory assumptions and proofs}
    \item[] Question: For each theoretical result, does the paper provide the full set of assumptions and a complete (and correct) proof?
    \item[] Answer: \answerYes{} 
    \item[] Justification: Seed Appendix \ref{app: proof}

    \item {\bf Experimental result reproducibility}
    \item[] Question: Does the paper fully disclose all the information needed to reproduce the main experimental results of the paper to the extent that it affects the main claims and/or conclusions of the paper (regardless of whether the code and data are provided or not)?
    \item[] Answer: \answerYes{} 
    \item[] Justification: Experimental details are given in Appendix \ref{app: experimental details}.

\item {\bf Open access to data and code}
    \item[] Question: Does the paper provide open access to the data and code, with sufficient instructions to faithfully reproduce the main experimental results, as described in supplemental material?
    \item[] Answer: \answerNo{} 
    \item[] Justification: Code will be released in the future.

\item {\bf Experimental setting/details}
    \item[] Question: Does the paper specify all the training and test details (e.g., data splits, hyperparameters, how they were chosen, type of optimizer, etc.) necessary to understand the results?
    \item[] Answer: \answerYes{} 
    \item[] Justification: Experimental settings are given in Section \ref{section: experiments} and Appendix \ref{app: experimental details}. Details are given in Appendix \ref{app: experimental details}.

\item {\bf Experiment statistical significance}
    \item[] Question: Does the paper report error bars suitably and correctly defined or other appropriate information about the statistical significance of the experiments?
    \item[] Answer: \answerYes{} 
    \item[] Justification: We use sufficient random seeds and provide standard deviation of the scores in Appendix \ref{app: full results}.

\item {\bf Experiments compute resources}
    \item[] Question: For each experiment, does the paper provide sufficient information on the computer resources (type of compute workers, memory, time of execution) needed to reproduce the experiments?
    \item[] Answer: \answerYes{} 
    \item[] Justification: Details of computation resources are given in Appendix \ref{app: experimental details}.
    
\item {\bf Code of ethics}
    \item[] Question: Does the research conducted in the paper conform, in every respect, with the NeurIPS Code of Ethics \url{https://neurips.cc/public/EthicsGuidelines}?
    \item[] Answer: \answerYes{} 
    \item[] Justification: This paper doesn't involve human participants, and the datasets in Section \ref{section: experiments} are common open-sourced datasets.

\item {\bf Broader impacts}
    \item[] Question: Does the paper discuss both potential positive societal impacts and negative societal impacts of the work performed?
    \item[] Answer: \answerYes{} 
    \item[] Justification: Potential positive impacts and negative impacts are discussed in Appendix \ref{Boarder Impacts}.
    
\item {\bf Safeguards}
    \item[] Question: Does the paper describe safeguards that have been put in place for responsible release of data or models that have a high risk for misuse (e.g., pretrained language models, image generators, or scraped datasets)?
    \item[] Answer: \answerNA{} 
    \item[] Justification: 

\item {\bf Licenses for existing assets}
    \item[] Question: Are the creators or original owners of assets (e.g., code, data, models), used in the paper, properly credited and are the license and terms of use explicitly mentioned and properly respected?
    \item[] Answer: \answerYes{} 
    \item[] Justification: All assets we use are properly credited, and the corresponding license are given in Appendix \ref{app: experimental details}.

\item {\bf New assets}
    \item[] Question: Are new assets introduced in the paper well documented and is the documentation provided alongside the assets?
    \item[] Answer: \answerNA{} 
    \item[] Justification: 

\item {\bf Crowdsourcing and research with human subjects}
    \item[] Question: For crowdsourcing experiments and research with human subjects, does the paper include the full text of instructions given to participants and screenshots, if applicable, as well as details about compensation (if any)? 
    \item[] Answer: \answerNA{} 
    \item[] Justification: 

\item {\bf Institutional review board (IRB) approvals or equivalent for research with human subjects}
    \item[] Question: Does the paper describe potential risks incurred by study participants, whether such risks were disclosed to the subjects, and whether Institutional Review Board (IRB) approvals (or an equivalent approval/review based on the requirements of your country or institution) were obtained?
    \item[] Answer: \answerNA{} 
    \item[] Justification: 

\item {\bf Declaration of LLM usage}
    \item[] Question: Does the paper describe the usage of LLMs if it is an important, original, or non-standard component of the core methods in this research? Note that if the LLM is used only for writing, editing, or formatting purposes and does not impact the core methodology, scientific rigorousness, or originality of the research, declaration is not required.
    \item[] Answer: \answerNA{} 
    \item[] Justification: 

\end{enumerate}

\end{document}